\documentclass[10pt,twocolumn,letterpaper]{article}

\usepackage{cvpr}              

\usepackage[accsupp]{axessibility} 

\usepackage{bm}             
\usepackage{booktabs}
\usepackage{multirow}
\usepackage[subtle]{savetrees} 
\usepackage{algorithm}
\usepackage{algorithmic}
\usepackage{amsthm}         
\usepackage{float}

\usepackage[pagebackref,breaklinks,colorlinks]{hyperref}

\usepackage[capitalize]{cleveref}
\crefname{section}{Sec.}{Secs.}
\Crefname{section}{Section}{Sections}
\Crefname{table}{Table}{Tables}
\crefname{table}{Tab.}{Tabs.}

\usepackage{cancel}
\usepackage[dvipsnames]{xcolor}

\DeclareMathOperator*{\argmax}{arg\,max}
\DeclareMathOperator*{\argmin}{arg\,min}

\DeclareMathOperator{\diag}{diag}

\DeclareMathOperator{\tr}{tr}

\DeclareMathOperator{\dist}{dist}

\DeclareMathOperator{\im}{Im}
\DeclareMathOperator{\rank}{rank}

\DeclareMathOperator{\FMS}{FMS}
\DeclareMathOperator{\TME}{TME}

\DeclareMathOperator{\STE}{STE}
\DeclareMathOperator{\ali}{align}

\DeclareMathOperator{\vecc}{vec}

\newtheorem{theorem}{Theorem}

\def\Dcal{{\mathcal{D}}}

\def\Xcal{{\mathcal{X}}}

\def\Rbb{{\mathbb{R}}}

\def\tbm{{\bm{t}}}
\def\ubm{{\bm{u}}}

\def\xbm{{\bm{x}}}

\def\Ibf{{\mathbf{I}}}

\def\Rbf{{\mathbf{R}}}
\def\Sbf{{\mathbf{S}}}
\def\Tbf{{\mathbf{T}}}
\def\Ubf{{\mathbf{U}}}

\def\Xbf{{\mathbf{X}}}

\def\Zbf{{\mathbf{Z}}}

\def\R{{\mathbb{R}}}

\def\bA{{\mathbf{A}}}

\def\bM{{\bf{M}}}
\def\bV{{\mathbf{V}}}
\def\bU{{\mathbf{U}}}
\def\bx{\mathbf{x}}
\def\bF{{\mathbf{F}}}
\def\b0{{\mathbf{0}}}
\def\bX{{\mathbf{X}}}
\def\bE{{\mathbf{E}}}
\def\bI{{\mathbf{I}}}
\def\bP{{\mathbf{P}}}

\def\bSigma{{\mathbf{\Sigma}}}

\def\calA{{\mathcal{A}}}

\def\calS{{\mathcal{S}}}
\def\calX{{\mathcal{X}}}

\def\paperID{17040} 
\def\confName{CVPR}
\def\confYear{2024}

\title{A Subspace-Constrained Tyler's Estimator\\ and its Applications to Structure from Motion \thanks{This work was supported by NSF DMS awards 2124913 and 2318926.
}}

\author{Feng Yu \thanks{Supplementary code: \href{https://github.com/alexfengg/STE}{https://github.com/alexfengg/STE}}\\
University of Minnesota\\
{\tt\small fyu@umn.edu}
\and
Teng Zhang\\
University of Central Florida\\
{\tt\small teng.zhang@ucf.edu}
\and
Gilad Lerman \thanks{Corresponding author.  All authors equally contributed. }\\ 
University of Minnesota\\
{\tt\small lerman@umn.edu}
}

\begin{document}

\maketitle

\begin{abstract}
We present the subspace-constrained Tyler's estimator (STE) designed for recovering a low-dimensional subspace within a dataset that may be highly corrupted with outliers. STE is a fusion of the Tyler's M-estimator (TME) and a variant of the fast median subspace. Our theoretical analysis suggests that, under a common inlier-outlier model, STE can effectively recover the underlying subspace, even when it contains a smaller fraction of inliers relative to other  methods in the field of robust subspace recovery. We apply STE in the context of 
Structure from Motion (SfM) in two ways: for robust estimation of the fundamental matrix and for the removal of outlying cameras, enhancing the robustness of the SfM pipeline. Numerical experiments confirm the state-of-the-art performance of our method in these applications. 
This research makes significant contributions to the field of robust subspace recovery, particularly in the context of computer vision and 3D reconstruction.
\end{abstract}

\section{Introduction}

In many applications, data has been collected in large quantities and dimensions. It is a common practice to represent such data within a low-dimensional subspace that preserves its essential information. Principal Component Analysis (PCA) is frequently employed to identify this subspace. However, PCA faces challenges when dealing with data contaminated by outliers. Consequently, the field of Robust Subspace Recovery (RSR) aims to develop a framework for outlier-robust PCA. RSR is particularly relevant to problems in computer vision, such as fundamental matrix estimation, which involves recovering a hidden subspace associated with ``good correspondence pairs'' among highly corrupted measurements.

Various algorithms have been proposed to address RSR, 
employing methods such as projection pursuit 
\cite{Ammann1993,choulakian06,1672644,Huber2009,kwak08,Li_85,robust_mccoy}, 
subspace energy minimization (in particular least absolute deviations and its relaxations) \cite{Ding+06,lerman2018fast,srebro2003weighted, lerman15reaper,zhang14gme, maunu19ggd,maunu22private,SteinhardtCV18}, robust covariance estimation \cite{zhang2016robust}, filtering-based methods \cite{Xu2010,Diakonikolas18,cherapanamjeri2017thresholding} and exhaustive subspace search methods \cite{fischler1981random,hardt2013algorithms}.
An in-depth exploration and comprehensive overview of robust subspace recovery and its diverse algorithms can be found in \cite{lerman2018overview}.

Methods based on robust covariance estimators, such as the Tyler's M-estimator (TME), offer additional useful information on the shape of the data within the subspace, similarly to PCA in the non-robust setting. They also offer maximum-likelihood interpretation, which is missing in many other methods. Application of the TME \cite{tyler1987distribution} to RSR has been shown to be successful on basic benchmarks~\cite{lerman2018overview,zhang2016robust}. Moreover, under a model of inliers in a general position on a subspace and outliers in general position in the complement of the subspace, TME was shown to recover the subspace within a desirable fraction of inliers \cite{zhang2016robust}. Below   
this fraction it was proved to be Small Set Expansion (SSE) hard to solve the RSR problem \cite{hardt2013algorithms}.

One may still succeed with solving the RSR problem with a computationally efficient algorithm when the fraction of inliers is lower than the one required by 
\cite{hardt2013algorithms}, considering a more restricted data model or violating other assumptions made in \cite{hardt2013algorithms}. For example, some special results in this direction are discussed in \cite{maunu2019robust}. Also, \cite{maunu19ggd} proposes the generalized haystack model of inliers and outliers to demonstrate the possibility of handling lower fractions of inliers by an RSR algorithm. This model extends the limited standard haystack model \cite{lerman15reaper}, where basic methods (such as PCA filtering) can easily work with low fractions of outliers. Nevertheless, it is unclear how practical the above theoretical ideas are for applied settings. 

One practical setting that requires a fraction of inliers significantly lower than the one stated in \cite{hardt2013algorithms} arises in the problem of robust fundamental (or essential) matrix estimation. 
The fundamental matrix encompasses the epipolar geometry of two views in stereo vision systems. It is typically computed using point correspondences between the two projected images. 
This computation requires finding an 8-dimensional subspace within a 9-dimensional ambient space. 
In this setting, the theoretical framework of \cite{hardt2013algorithms} requires that the fraction of inliers be at least $8/9\approx 88.9\%$, which is clearly unreasonable to require.  

To date, the RANdom Sample Consensus (RANSAC) method~\cite{fischler1981random} is the only RSR method that has been highly successful in addressing this nontrivial scenario, gaining widespread popularity in computer vision. RANSAC is an iterative method that
randomly selects minimal subsets of the data and fits models, in particular subspaces, to identify the best consensus set, that is, the set in most agreement with the hypothesized model.  
There are numerous approaches proposed to improve RANSAC, especially for this particular application, including locally optimized RANSAC (LO-RANSAC, \cite{chum2003locally}), maximum likelihood estimator RANSAC (MLESAC) \cite{torr2000mlesac}), degeneracy-check enabled RANSAC (DEGENSAC) \cite{chum2005two}) and M-estimator guided RANSAC (MAGSAC) \cite{barath2019magsac}). 
A near recovery theory for a variant of RANSAC under
some assumptions on the outliers was suggested in \cite{maunu2019robust}. 
Nevertheless, in general, RANSAC is rather slow and its application to higher-dimensional problems is intractable.  

This work introduces a novel RSR algorithm that is guaranteed to robustly handle a lower fraction of outliers than the theoretical threshold proposed by \cite{hardt2013algorithms}, under special settings.
Our basic idea is to adapt Tyler's M-Estimator to utilize the information of the underlying $d$-dimensional subspace, while avoiding estimation of the full covariance. By using less degrees of freedom we obtain a more accurate subspace estimator than the one obtained by TME with improved computational complexity. 
We show that STE is a fusion of the Tyler's M-estimator (TME) and a variant of the fast median subspace (FMS)~\cite{lerman2018fast} that aims to minimize a subspace-based $\ell_0$ energy.

Our theory shows that our proposed subspace-constrained Tyler's estimator (STE) algorithm can effectively recover the underlying subspace, even when it contains a smaller fraction of inliers relative to other methods. We obtain this nontrivial achievement first in a generic setting, where we establish when an initial estimator for STE is sufficiently well-conditioned to guarantee the desired robustness of STE. We then assume the asymptotic generalized haystack model and show that under this model, TME itself is a well-conditioned initial estimator for STE, and that unlike TME, STE with this initialization can deal with a lower fraction of inliers than the theoretical threshold specified in \cite{hardt2013algorithms}.

We demonstrate competitive performance in robust fundamental matrix estimation, relying solely on subspace information without additional methods for handling degenerate scenarios, in contrast to
\cite{chum2005two,frahm2006ransac,raguram2012usac}. We also propose a potential application of RSR for removing bad cameras in order to enhance the SfM pipeline and show competitive performance of STE. This is a completely new idea and it may require additional exploration to make it practical. Nevertheless, it offers a very different testbed where $N=D$ is very large and RANSAC is generally intractable.

The rest of the paper is organized as follows: \S\ref{sec:STE} introduces the STE framework, \S\ref{sec:theory} establishes theoretical guarantees of STE, \S\ref{sec:applications} applies STE to two different problems in SfM, demonstrating its competitive performance relative to existing algorithms, and \S\ref{sec:conclusions} provides conclusions and future directions.

\section{The STE Algorithm}\label{sec:STE}

We present our proposed STE. 
We review basic notation in \S\ref{sec:notation} and Tyler's original estimator in \S\ref{sec:tme}. We describe our method in \S\ref{sec:ste_describe}, 
its computational complexity in \S\ref{sec:comp_complexity}, its  algorithmic choices in \S\ref{sec:ste_parameters} and an interpretation for it as a fusion of TME and FMS with $p=0$ in \S\ref{sec:ste_fusion}.

\subsection{Notation}
\label{sec:notation}
We use bold upper and lower case letters for matrices and column vectors, respectively. Let $\bI_k$ denote the identity matrix in $\mathbb{R}^{k \times k}$, where if $k$ is obvious we just write $\bI$. For a matrix $\bA$, we denote by $\tr(\bA)$ and $\im(\bA)$ the trace and image (i.e., column space) of $\bA$. 
We denote by $S_{+}(D)$ and $S_{++}(D)$ the sets of positive semidefinite and definite matrices in $\mathbb{R}^{D \times D}$, respectively.
We denote by $O(D,d)$ the set of semi-orthogonal ${D \times d}$ matrices, i.e., $\bU \in O(D,d)$ if and only if $\bU \in \mathbb{R}^{D \times d}$ and $\bU^\top \bU = \bI_{d}$.  
We refer to linear $d$-dimensional subspaces as $d$-subspaces.
For a $d$-subspace $L$, we denote by $\bP_L$ the $D\times D$ matrix representing the orthogonal projector onto $L$. 
We also arbitrarily fix $\bU_L$ in $O(D,d)$ such that $\bU_L \bU_L^\top = \bP_L$ (such $\bU_L$ is determined up to right multiplication by an orthogonal matrix in $O(d,d)$). 
Throughout the paper,  $\calX = \{\xbm_i\}_{i=1}^N\subset\Rbb^D$ is assumed to be a given centered dataset, that is, $\sum_{i=1}^N \xbm_i = \b0$. 

\subsection{Tyler's Estimator and its Application to RSR}
\label{sec:tme}
Tyler's M-estimator (TME)~\cite{tyler1987distribution} robustly estimates the covariance $\bSigma^*$ of the dataset  
$\calX=\{\xbm_i\}_{i=1}^N\subset\Rbb^D$
by minimizing
\begin{align}\label{eq:TME}
    \frac{D}{N} \, \sum_{i=1}^N\log(\xbm_i^\top \bSigma^{-1}\xbm_i)+\log\det(\bSigma) 
\end{align}
over $\bSigma\in S_{++}(D)$ such that  $\tr(\bSigma)=1$.  
The cost function in \eqref{eq:TME} can be motivated by writing the maximum likelihood of the multivariate $t$-distribution and letting its degrees of freedom parameter, $\nu$, approach zero~\cite{maronna2014robust}. 
This cost function is invariant to dilations of $\bSigma$, and the constraint on $\tr(\bSigma)$, whose value can be arbitrarily chosen, fixes a scale.  
TME also applies to scenarios where the covariance matrix does not exist. In such cases, TME estimates the shape (or scatter matrix) of the distribution, which is defined up to an arbitrary scale. More direct interpretations of TME as a maximum likelihood estimator can be found in \cite{tyler1987angularGaussian,Frahm2010generalizedTME}. When $D$ is fixed and $N$ approaches infinity, TME is the “most robust” estimator of the shape matrix for data i.i.d.~sampled from a continuous elliptical distribution \cite{tyler1987distribution} in a minimax sense, that is, as a minimizer of the maximal variance.

Tyler \cite{tyler1987distribution} proposed the  following iterative formula for computing TME:  
\begin{equation}
\nonumber
    \bSigma^{(k)} = \sum_{i=1}^N\frac{ \xbm_i \, \xbm_i^\top}{\xbm_i^\top (\bSigma^{(k-1)})^{-1}\xbm_i}/\tr\left( \sum_{i=1}^N\frac{\xbm_i \, \xbm_i^\top}{\xbm_i^\top (\bSigma^{(k-1)})^{-1}\xbm_i} \right).
\end{equation}
Kent and Tyler \cite{Kent_Tyler88} proved that if any $d$-subspace of $\mathbb{R}^D$, where $1 \leq d \leq D-1$, contains fewer than $Nd/D$ data points, then the above iterative procedure converges to TME. 
Linear rate of convergence was proved for the regularized TME in \cite{goes20} and for TME in \cite{franks2020TMElinearrate}. 

One can apply the TME estimator to solve the RSR problem with a given dimension $d$ by forming the subspace spanned by the top $d$ eigenvectors of TME. 
Zhang \cite{zhang2016robust} proved that as long as there are more than $Nd/D$ inliers lying on a subspace, and the projected coordinates of these inliers on the $d$-subspace and the projected coordinates of the outliers on the $(D-d)$-dimensional orthogonal complement of the subspace are in general position, then TME recovers this subspace. 
Zhang \cite{zhang2016robust} also showed that in this setting the above iterative formula converges
(note that the condition of convergence in \cite{Kent_Tyler88} does not apply in this case). 
The above lower bound of $Nd/D$ on the number of inliers coincides with the
general bound for the noiseless RSR problem, beyond which the problem becomes Small Set-Expansion (SSE) hard \cite{hardt2013algorithms}.

Numerical experiments in \cite{zhang2016robust} and \cite{lerman2018overview} 
indicated state-of-the-art accuracy of TME compared to other RSR algorithms in various settings. The computational complexity of TME is of order $O(K(N D^2 +D^3))$, where $K$ is the number of iterations. On the other hand,  the cost of faster RSR algorithms is of order $O(KNDd)$ \cite{lerman2018fast,maunu19ggd,lerman2018overview}.

\subsection{Motivation and Formulation of STE}
\label{sec:ste_describe}
We aim to use more cleverly the $d$-subspace information within the TME framework to form an RSR algorithm, instead of first estimating the full covariance. By using less degrees of freedom we can obtain a more accurate subspace estimator, especially when the fraction of outliers can be large. Furthermore, our idea allows us to improve the computational cost to become state-of-the-art for high-dimensional settings.  

Many RSR algorithms can be formulated as minimizing a best orthogonal projector onto a $d$-subspace \cite{zhang14gme,lerman15reaper,lerman2018fast,maunu19ggd,lerman2018overview}. We are going to do something similar, but unlike using an orthogonal projector, we will still use information from TME to get the shape of the data on the projected subspace. We will make the rest of the eigenvalues (i.e., bottom $D-d$ ones) equal and shrink them by a parameter $0<\gamma<1$. We thus  use a regularized version of a reduced-dimension covariance matrix. This parameter $\gamma$ plays a role in our theoretical estimates. 
Making $\gamma$ smaller helps with better subspace recovery, whereas making $\gamma$ bigger enhances the well-conditioning of the estimator. 

Following these basic ideas, we formulate our method, STE. For simplicity, we utilize covariance matrices and their inverses. Since these covariance matrices are essentially $d$-dimensional and include an additional simple regularizing component, our overall computations can be expressed in terms of the computation of the top $d$ singular values of an $N \times D$ matrix (see \S\ref{sec:comp_complexity}).

At iteration $k$ we follow a similar step to that of TME: 
$$\Zbf^{(k)} := \sum_{i=1}^N\xbm_i\xbm_i^\top/(\xbm_i^\top(\bSigma^{(k-1)})^{-1}\xbm_i).$$
We compute the eigenvalues 
 $\{\sigma_i\}_{i=1}^D$ of $\Zbf^{(k)}$ and replace each of the bottom $(D-d)$  of them with $\gamma \cdot \sigma_{d+1,D}$, where
\begin{align}\label{eq:last_eig_scaling}
    \sigma_{d+1,D} :=  \frac{1}{D-d}\sum_{i=d+1}^D\sigma_i.
\end{align}
We also compute the eigenvectors of $\Zbf^{(k)}$ and form the matrix $\bSigma^{(k)}$ with the same eigenvectors as those of 
$\Zbf^{(k)}$ and the modified eigenvalues, scaled to have trace 1. We iteratively repeat this procedure until the two estimators are sufficiently close. \Cref{alg:ste} summarizes this procedure. Note that it is invariant to  scaling of the data, similarly to TME.

\begin{algorithm}
  \caption{STE: Subspace-Constrained Tyler's Estimator}
  \label{alg:ste}
  \begin{algorithmic}[1]
    \STATE \textbf{Input:} $\Xbf=[\xbm_1,\ldots,\xbm_N]\in\Rbb^{D\times N}$: centered data matrix, $d$: subspace dimension, $K$: maximum number of iterations, $\tau,\gamma$: parameters.
    \STATE \textbf{Output:} $L$: $d$-subspace in $\Rbb^D$
    \STATE $\bSigma^{(0)}=\Ibf_D/D$
    \FOR{$k=1,2,\ldots$}
        \STATE 
        $\Zbf^{(k)}\leftarrow
\sum_{i=1}^N\xbm_i\xbm_i^\top/(\xbm_i^\top(\bSigma^{(k-1)})^{-1}\xbm_i)$
        \STATE 
        $[\Ubf^{(k)},\Sbf^{(k)},\Ubf^{(k)}]\leftarrow\text{EVD}(\Zbf^{(k)})$
        \STATE 
        $\sigma_i \leftarrow [\Sbf^{(k)}]_{ii}$
        and $ \sigma_{d+1,D} \leftarrow \sum_{i=d+1}^D {\sigma_i}/{(D-d)}$ 
        \STATE  
$\widetilde{\Sbf}^{(k)}\leftarrow\diag( \sigma_1,\ldots, \sigma_d, \gamma \cdot  \sigma_{d+1,D},\ldots, \gamma \cdot \sigma_{d+1,D})$, 
        \\
        \STATE  $\bSigma^{(k)}\leftarrow\Ubf^{(k)}\widetilde{\Sbf}^{(k)}(\Ubf^{(k)})^\top/ \tr\big(\Ubf^{(k)}\widetilde{\Sbf}^{(k)}(\Ubf^{(k)})^\top\big)$
        
        \STATE Stop if $k\geq K$ or $\|\bSigma^{(k)}-\bSigma^{(k-1)}\|_F<\tau$ 
    \ENDFOR
    \STATE $L$ = Span of the first $d$ columns of $\Ubf^{(k)}$
  \end{algorithmic}
\end{algorithm}
\subsection{Computational Complexity}
\label{sec:comp_complexity}

Setting ${w}_i^{(k)} = (\xbm_i^\top(\bSigma^{(k-1)})^{-1}\xbm_i)^{-1}$, we can express $\Zbf^{(k)}$ as    $\Zbf^{(k)}=\widetilde{\Xbf}\widetilde{\Xbf}^\top$,  where $\widetilde{\Xbf} = [({w}_1^{(k)})^{1/2}\xbm_1,\ldots,({w}_N^{(k)})^{1/2}\xbm_N]$.
With some abuse of notation we denote by $\sigma_1,\ldots,\sigma_D$ the eigenvalues of $\bSigma^{(k-1)}$ (and not $\bSigma^{(k)}$). 
Since they are scaled to have trace 1,  $\sigma_{d+1,D}=(1-\sum_{j=1}^d\sigma_j)/(D-d)$. We thus only need the top $d$ eigenvectors and top $d$ eigenvalues of $\bSigma^{(k-1)}$ to update $\widetilde{w}_i^{(k)}$.  Therefore, the complexity of STE can be of order $O(KNDd)$ if a special fast algorithm is utilized for computing only the top $d$ eigenvectors.  

\subsection{Implementation Details}
\label{sec:ste_parameters}

STE depends on the parameters $K$, $\tau$ and $\gamma$ and the initialization of $\bSigma^{(0)}$. 
The first two parameters are rather standard in iterative procedures and do not raise any concern. 

Our theory sheds some light on possible choices of $\gamma$ and in particular it indicates that the algorithm can be more sensitive to choices of $\gamma$ when the quantity defined later in \eqref{eq:dssnr} is relatively small. In this case, it may be beneficial to try several values of $\gamma$.  
We propose here a constructive way of doing it. We first form a sequence of $0<\gamma\leq 1$,
e.g., $\gamma_k = 1/k$, $k=1,\ldots,m$. 
In order to determine the best choice of $\gamma$, we compute the distance of each data point $\xbm$ to each subspace $L_k$, corresponding to the choice of $\gamma_k$, where  $\dist(\xbm,L_k)=\|\xbm-\bP_{L_k}\xbm\|$. We set set a threshold $\zeta$, 
obtained by the median among all points and all subspaces and for each subspace, $L_k$, we count the number of the inliers with distance below this threshold. The best $\gamma_k$ is determined according to the subspace yielding the largest number of inliers. We describe this procedure in \Cref{alg:ste-lambda}. 

For simplicity, we  initialize with $\bSigma^{(0)}=\Ibf_D/D$ and note that with this choice $\bSigma^{(1)}$
reflects the trimmed covariance matrix and thus reflects the PCA subspace. One can also initialize with TME or other subspaces (see \S\ref{sec:theory} 
where the theory of STE is discussed). One can further try several initialization (with possible random components) and use a strategy similar to 
\Cref{alg:ste-lambda} to choose the best one. 

At last, we remark that when computing 
$\Zbf^{(k)}$ we want to ensure that $\xbm_i^\top(\bSigma^{(k-1)})^{-1}\xbm_i$ cannot be zero and we thus add the arbitrarily small number  $10^{-15}$ to this value.

\begin{algorithm}
  \caption{Estimating best $\gamma$ for STE}
  \label{alg:ste-lambda}
  \begin{algorithmic}[1]
    \STATE \textbf{Input:} $\Xbf=[\xbm_1,\ldots,\xbm_N]\in\Rbb^{D\times N}$: centered data matrix, $d$: subspace dimension, $\{\gamma_1,\ldots,\gamma_m\}$: a set of pre-selected $\gamma$'s.
    \STATE \textbf{Output:} $\gamma^*$: optimal $\gamma$ among $\{\gamma_1,\ldots,\gamma_m\}$
    \FOR{$j=1,2,\ldots,m$}
        \STATE $L^{(j)}\leftarrow \STE(\Xbf,d,\gamma_j)$
        \STATE $\Dcal^{(j)}\leftarrow\{\dist(\xbm_i,L^{(j)})\mid\xbm_i\in\Xcal\}$. 
    \ENDFOR
    \STATE Set $\zeta=$median(\{$\Dcal^{(1)}$,\ldots,$\Dcal^{(m)}$\})
    \STATE $j^*=\argmax_{1\leq j\leq m}|\Dcal^{(j)}<\zeta|$
    \STATE $\gamma^* = \gamma_{j^*}$
  \end{algorithmic}
\end{algorithm}
\vspace{-.1in}
\subsection{STE fuses TME and a Variant of FMS}
\label{sec:ste_fusion}
STE is formally similar to both TME and FMS. Indeed, at each iteration these algorithms essentially compute
    $\bSigma^{(k+1)}  = \sum_{i=1}^N
    w_i \xbm_i\xbm_i^\top$,
where $w_i \equiv w_i\hspace{-.03in}\left(\bSigma^{(k)}\right)$.
We summarize the formal weights for FMS (with any choice of $p$ for minimizing an $\ell_p$ energy in \cite{lerman2018fast}), TME and STE. We ignore an additional scaling constant for TME and STE, obtained by dividing $w_i \xbm_i\xbm_i^\top$ above by its trace, and a regularization parameter $\delta$ for FMS.   
We express these formulas using the eigenvalues 
$\sigma_1,\ldots,\sigma_D$ and eigenvectors $\ubm_1,\ldots,\ubm_D$ of the weighted sample covariance,  $\sum_{i=1}^N w_i \xbm_i\xbm_i^\top$ for each method and $\beta : = \gamma \cdot \sigma_{d+1,D}$ (see \eqref{eq:last_eig_scaling}) as follows:
\begin{align*}
    w_i^{\FMS} & = \frac{1}{\left(\sum_{j=d+1}^D(\xbm_i^\top\ubm_j)^2\right)^{(2-p)/2}}, \\
    w_i^{\TME} & = 
    \frac{1}{\sum_{j=1}^D\sigma_j^{-1}(\xbm_i^\top\ubm_j)^2}, \\
    w_i^{\STE} & = \frac{1}{\sum_{j=1}^d\sigma_j^{-1}(\xbm_i^\top\ubm_j)^2+\beta^{-1} \sum_{j=d+1}^D(\xbm_i^\top\ubm_j)^2}.
\end{align*}

These weights aim to mitigate the impact of outliers in different ways. 
For FMS, $\sum_{j=d+1}^D(\xbm_i^\top\ubm_j)^2$ is the squared distance of a data point $\xbm_i$ to the subspace $L$.
Thus for $p<2$, $w_i^{\FMS}$ is smaller for ``subspace-outliers'', where the robustness to such outliers increases when  $p \geq 0$ decreases. 

The weights of TME are inversely proportional to the squared Mahalanobis distance of $\xbm_i$ to the empirical distribution. They mitigate the effect of ``covariance-outliers''. If the dataset is concentrated on a $k$-subspace where $k<d$, then TME can provide smaller weights to  points lying away from this subspace, unlike FMS that does not distinguish between points within the larger $d$-subspace. 

We note that the weights of STE fuse the above two weights. Within a $d$-subspace, they use the shape of the data. They can thus avoid outliers within this $d$-subspace. Within the orthogonal component of this subspace, they use a term proportional to that of FMS with $p=0$. We remark that such $\ell_0$ minimization has a clear interpretation for RSR, though is generally hard to guarantee. Indeed,  \cite{lerman2018fast} has no guarantees for FMS with $p=0$.  
It can also yield unwanted spurious stationary points \cite{lermanzhang2014lpsubspaces}.

\section{Theory}
\label{sec:theory}
We review a theoretical guarantee for STE, whose proof is given in \cite{lerman2024theoretical}. It requires some conditions and we verify they hold with high probability under the asymptotic generalized haystack model. We assume a noiseless inliers-outliers RSR model. Let $L_*$ denote the underlying $d$-subspace in $\mathbb{R}^D$, $\calX_{in}=\calX \cap L_*$ and  $\calX_{out}=\calX\setminus \calX_{in}$ be the set of inliers and outliers, respectively, and  $n_1=|\calX_{in}|$ and $n_0=|\calX_{out}|$ be the number of inliers and outliers. 
Our first assumption is a mild one on how well-conditioned the inliers are in $L_*$ (compare e.g., other assumptions in \cite{lerman2018overview,maunu2019robust}).

\noindent
\textbf{Assumption 1:} Any $k$-subspace of $L_*$, $1 \leq k \leq d$, contains at most $n_1k/d$ points.

\textbf{Motivation for Assumption 2:}
The ratio of inliers per outliers, $n_1/n_0$, in RSR is often referred to as the SNR (signal-to-noise ratio)~\cite{maunu19ggd,lerman2018overview,maunu2019robust}. The smaller it is, the best the subspace recovery is. We define the dimension-scaled SNR (DS-SNR) as the SNR obtained when scaling $n_1$ and $n_0$ by their respective dimensions (of $L_*$ and $L_*^{\perp}$):
\begin{equation}
\label{eq:dssnr}
    \text{DS-SNR}:= \frac{n_1/d}{n_0/(D-d)}.
\end{equation}
Zhang \cite{zhang2016robust} showed that exact recovery by TME is guaranteed whenever DS-SNR$>1$ (assuming general position assumptions on the inliers and outliers) and Hardt and Moitra \cite{hardt2013algorithms} showed that when considering general datasets with general position assumptions on the inliers and outliers, the RSR problem is SSE hard if the DS-SNR is lower than $1$.
We aim to show that under the following weaker generic condition, STE can obtain exact recovery  with DS-SNR, strictly lower than 1. \\
\noindent
\textbf{Assumption 2:} 
$\text{DS-SNR}> \gamma$, where $\gamma<1$ is the STE parameter.

Our last assumption requires a  sufficiently good initialization for STE, but also implicitly involves additional hidden assumptions on the inliers and outliers. This is expected, since Assumption 1 does not require anything from the outliers and also has a very weak requirement from the inliers. To formulate the new assumption we define below some some basic condition numbers for good initialization (which are more complicated than the one for initialization by PCA suggested by  \cite{maunu19ggd} and \cite{maunu2019robust}) and also quantities similar to the ones used to guarantee landscape stability in the theory of RSR \cite{lerman15reaper,zhang14gme,maunu19ggd,lerman2018overview}.

\textbf{Definitions required for Assumption 3:} 
Recall that $\bSigma^{(0)}$ denotes  the initial value in \Cref{alg:ste}, and denote
\[
\bSigma^{(0)}_{L_1,L_2}=\bU_{L_1}^\top\bSigma^{(0)}\bU_{L_2}. 
\]
We define the following  condition number
\[
\kappa_1=\frac{\sigma_d\Big(\bSigma^{(0)}_{L_*,L_*}-\bSigma^{(0)}_{L_*,L_*^\perp}\bSigma^{(0)\,-1}_{L_*^\perp,L_*^\perp}\bSigma^{(0)}_{L_*^\perp,L_*}\Big)}{\sigma_1\Big(\bSigma^{(0)}_{L_*^\perp,L_*^\perp}\Big)}.
\]
To get a better intuition to this primary quantity of Assumption 3, we first express the initial estimator $\bSigma^{(0)}$, using basis vectors for $L_*$ and $L_*^\perp$, as a $2\times 2$ block matrix 
\[
\begin{pmatrix}
\bSigma^{(0)}_{L_*,L_*} & \bSigma^{(0)}_{L_*,L_*^\perp} \\
\bSigma^{(0)}_{L_*^\perp,L_*} & \bSigma^{(0)}_{L_*^\perp,L_*^\perp} 
\end{pmatrix}.
\]
Defining $\bSigma'=\bSigma^{(0)}_{L_*,L_*^\perp}\bSigma^{(0)\,-1}_{L_*^\perp,L_*^\perp}\bSigma^{(0)}_{L_*^\perp,L_*}$, we decompose this block matrix as \[
\begin{pmatrix}
\bSigma' & \bSigma^{(0)}_{L_*,L_*^\perp} \\
\bSigma^{(0)}_{L_*^\perp,L_*} & \bSigma^{(0)}_{L_*^\perp,L_*^\perp}, 
\end{pmatrix}
+
\begin{pmatrix}
\bSigma^{(0)}_{L_*,L_*}-\bSigma' & 0 \\
0 & 0
\end{pmatrix}.
\]
We note that the numerator of $\kappa_1$ is the $d$-th eigenvalue of the second matrix in the above sum. We show in \cite{lerman2024theoretical} that this eigenvalue is positive if $\bSigma^{(0)}$ is positive definite, which can be easily enforced.
The condition number is thus the ratio between the smallest positive eigenvalue
of the second matrix of the sum and the largest eigenvalue of the component of the first matrix associated with $L_*^\perp$. 
Therefore, $\kappa_1$ expresses a ratio between a quantifier of a $d$-dimensional component of $\bSigma^{(0)}$, associated with $L_*$, and a quantifier of the projection onto $L_*^\perp$ of a full rank component of $\bSigma^{(0)}$.

We also define $\bSigma_{in,*}$
as the TME solution to the set of the projected inliers $\{\bU_{L^*}\xbm\mid\xbm\in\Xcal_{in}\}\subset\Rbb^d$ and
the following two condition numbers
\begin{equation*}
\kappa_2=\frac{\sigma_1\Big(\bSigma^{(0)}_{L_*^\perp,L_*^\perp}\Big)}{\sigma_D(\bSigma^{(0)})} \ \text{ and } \ \kappa_{in}=\frac{\sigma_1(\bSigma_{in,*})}{\sigma_d(\bSigma_{in,*})}.
\end{equation*}
We note that $\kappa_{in}$ is analogous to the condition number in (25) of \cite{maunu2019robust}, where we replace the sample covariance by the TME estimator.
An analog to the alignment of outliers statistic \cite{lerman15reaper,maunu19ggd} for STE is 
\begin{equation*}
\calA=\Big\|\sum_{\bx\in\calX_{out}}\frac{\bx\bx^\top}{\|\bU_{L_*^{\perp}}\bx\|^2}\Big\|.     
\end{equation*}
An analog to the stability statistic \cite{lerman15reaper,maunu19ggd} for STE 
is
\begin{equation*}
\calS=\sigma_{d+1,D}\Big(\sum_{\bx\in\calX}\frac{\bx\bx^\top}{\|\bx\|^2}\Big),    
\end{equation*}
where $\sigma_{d+1,D}(\bX)$ was defined in \eqref{eq:last_eig_scaling}.

\noindent
\textbf{Assumption 3:} There exists $C =C(\gamma,\text{DS-SNR})>0$ such that 
\begin{equation}\label{eq:kappa1}
\kappa_1\geq C \, \frac{d\,\,\kappa_{in}\,\calA}{n_1}\left(\kappa_{in}+\frac{\calA}{\frac{n_1}{d}-\gamma\frac{n_0}{D-d}}+\frac{\kappa_2\calA}{\gamma \calS}(1+\kappa_{in})\right).
\end{equation}
The exact technical requirement on $C$ is specified in \cite{lerman2024theoretical}.
In general, the larger the RHS of \eqref{eq:kappa1}, the more restricted the choice of $\bSigma^{(0)}$ is. In particular, when $\kappa_1=\infty$, the definition of $\kappa_1$ implies that $\im(\bSigma^{(0)})=L_*$, so the subspace is already recovered by the initial estimate. Therefore, reducing the lower bound of $\kappa_1$ may allow some flexibility, so a marginally suboptimal initialization could still work out. In \cite{lerman2024theoretical}, we show that under the asymptotic generalized haystack model, Assumption 3 can be interpreted as an upper bound on the largest principal angle between the initial and ground truth subspaces.



\textbf{Generic Theory:} The next theorem suggests that under assumptions 1-3, STE nicely converges to an estimator that recovers $L_*$.
The main significance of this theory is that its assumptions can allow DS-SNR lower than 1 for special instances of datasets (for which the assumptions hold), unlike the general recovery theories of~\cite{hardt2013algorithms} and \cite{zhang2016robust}. 

\begin{theorem}\label{thm:main}
Under assumptions 1-3,  the sequence $\bSigma^{(k)}$ generated by STE converges to $\bU_{L_*}\bSigma_{in,*}\bU_{L_*}^\top$, the TME solution for the set of inliers $\calX_{in}$. In addition, let $L^{(k)}$ be the subspace spanned by the top $d$ eigenvectors of $\bSigma^{(k)}$, then the angle between $L^{(k)}$ and $L_*$, $\angle(L^{(k)},L_*)=\cos^{-1}(\|\bU_{L^{(k)}}^\top\bU_{L_*}\|)$, converges $r$-linearly to zero.
\end{theorem}
We discuss insights of this theory on choices of the algorithms and further verify the above stated advantage of STE over TME assuming a common probabilistic model.

 \textbf{Choice of $\gamma$ for subspace recovery:} In order to avoid too large lower bound for $\kappa_1$ in \eqref{eq:kappa1}, which we motivated above, it is good to find $\epsilon_1$ and $\epsilon_2 >0$, such that 
$\gamma$ lies in $(\epsilon_1,$DS-SNR$-\epsilon_2)$
(to notice this, observe the terms involving $\gamma$ in the denominators of the last two additive terms in \eqref{eq:kappa1}). We thus note that if the DS-SNR is expected to be sufficiently larger than 1, we can use, e.g., $\gamma = 0.5$, but when the DS-SNR can be close to 1 or lower (e.g., in fundamental matrix estimation), it is advisable to choose small values of $\gamma$ according to \Cref{alg:ste-lambda} and their sizes may depend on the expected value of the DS-SNR.

\textbf{Possible ways of Initialization:} 
If one expects an initial estimated subspace $\hat{L}$ to have a sufficiently small angle $\theta$ with $L_*$, where  $\theta=\angle(\hat{L},L_*)$, then for $\bSigma^{(0)}:=\Pi_{\hat{L}}+\epsilon \bI$ it can be shown that   $\kappa_1>O(1/(\epsilon+\theta))$ and $\kappa_2<O(1+\frac{\theta}{\epsilon})$. Thus one may use a trusted RSR method, e.g., FMS. As discussed in \S\ref{sec:ste_parameters}, the choice $\bSigma^{(0)} = \bI$ (or a scaled version of it) corresponds to $\hat{L}$ being the PCA subspace (obtained at iteration 1). Also, using the TME solution for $\bSigma^{(0)}$ corresponds to using the TME subspace as $\hat{L}$.

\textbf{Theory under a probabilistic model:}
We show that under a common probabilistic model, the assumptions of Theorem \ref{thm:main}, where $\bSigma^{(0)}$ is obtained by TME, hold. Moreover, we show that STE (initialized by TME) can recover the correct subspace in situations with DS-SNR$< 1$, whereas TME cannot recover the underlying subspace in such cases.  
We follow \cite{maunu19ggd} and study the Generalized Haystack Model, though for simplicity, we assume Gaussian instead of sub-Gaussian distributions and an asymptotic setting. We assume $n_1$ inliers i.i.d.~sampled from a Gaussian distribution $N(0,\bSigma^{(in)}/d)$, where $\bSigma^{(in)} \in S_+(D)$ and $L_*=\im(\bSigma^{(in)})$ (so $\bSigma^{(in)}$ has $d$ nonzero eigenvalues), and $n_0$ outliers are i.i.d.~sampled from a Gaussian distribution $N(0,\bSigma^{(out)}/D)$, where  
$\bSigma^{(out)}/D \in S_{++}(D)$. 
We define the following condition numbers of inliers (in $L_*$) and outliers:
\[\kappa_{in}=\frac{\sigma_1(\bSigma^{(in)})}{\sigma_d(\bSigma^{(in)})}\,\,\ \ \text{ and } \ \kappa_{out}=\frac{\sigma_1(\bSigma^{(out)})}{\sigma_D(\bSigma^{(out)})}\,.\]

Clearly, Assumption 1 holds under this model, and Assumption 2 constrains some of its parameters. Our next theorem shows that 
Assumption 3 holds under this model when the initial estimate $\bSigma^{(out)}$ for STE is obtained by TME. It also shows that in this case STE can solve the RSR problem even when DS-SNR$<1$, unlike TME. For simplicity, we formulate the theory for the asymptotic case, where $N \rightarrow\infty$ and the theorem holds almost surely. It is possible to formulate it for a very large $N$ with high probability, but it requires stating complicated constants depending on various parameters. 

\begin{theorem}\label{thm:haystack}
Assume data generated from the above generalized haystack model. Assume further that 
for $0<\mu<1$, which can be arbitrarily small, $d<(1-\mu)D-2$. 
Then, for any chosen $0<c_0<1$, which is a lower bound for $\gamma$, there exists $\eta:=\eta(\kappa_{in},\kappa_{out},c_0,\mu)<1$ such that if DS-SNR$\geq \eta$ 
and $\bSigma^{(0)}$ is obtained by TME, then Assumption 3 for $\bSigma^{(0)}$ is satisfied with  $c_0<\gamma<\eta-c_0$ almost surely as $N\rightarrow\infty$. 
Consequently, the output of the STE algorithm, initialized by TME and with the choice of  $c_0<\gamma<\eta-c_0$, recovers $L_*$. On the other hand, if $\bSigma^{(out)}_{L_*,L_*^\perp}\neq 0$ and DS-SNR$ < 1$, then the top $d$ eigenvectors of TME do not recover $L_*$.
\end{theorem}

There are three different regimes that the theorem covers. When DS-SNR$\geq 1$, both TME+STE (i.e., STE initialized by TME) and TME solve the RSR problem. When $\eta \leq$DS-SNR$<1$, TME+STE solves the RSR problem and TME generally fails. When 
$\gamma \leq$DS-SNR$<\eta$, TME+STE might also fail, but STE with extremely good initialization (that satisfies Assumption 3) can still solve the problem. 

To get a basic idea of the dependence of $\eta$ on its parameters, we remark that $\eta \rightarrow 1$ if either $c_0 \rightarrow 0$, $\kappa_{in} \rightarrow \infty$,  $\kappa_{out} \rightarrow \infty$ or $\mu \rightarrow 0$, where the parameter $\mu$ is somewhat artificial and might be removed with a tighter proof. 
Therefore, successful performance of TME+STE requires a DS-SNR that is close to $1$ when $\gamma$ is close to either $0$ or $\eta$ (so that $c_0$ is very small) or when either the inlier or outlier distribution is highly non-symmetric, that is, either $\kappa_{in}$ or $\kappa_{out}$ is large.

\begin{figure*}[htbp]
    \centering
    \includegraphics[width=0.9\linewidth]{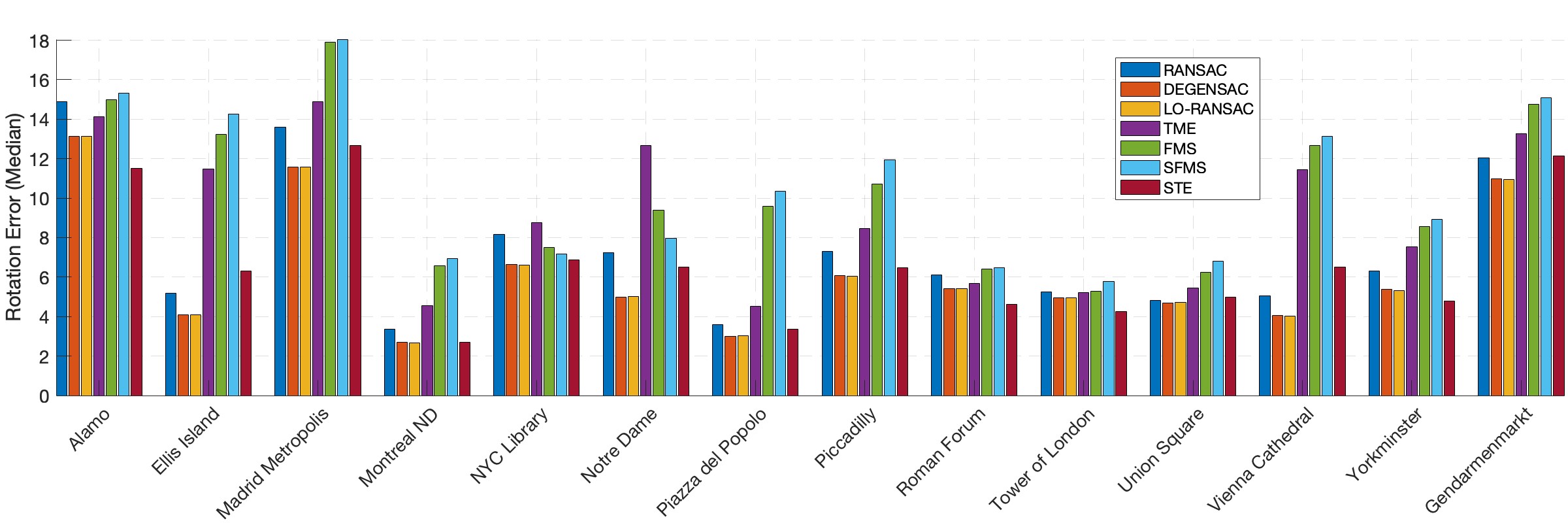}
    \caption{Median (relative) rotation errors obtained by seven algorithms for the 14 datasets of Photo Tourism.}
    \label{fig:fund_plot}
\end{figure*}

\section{Applications to Structure from Motion}\label{sec:applications}

We apply STE to problems relevant to SfM: 
robust estimation of fundamental matrices (see \S\ref{sec:robust_fm}), and initial screening of undesirable cameras (see \S\ref{sec:sfm}).

\subsection{Robust Fundamental Matrix Estimation}\label{sec:robust_fm}
Fundamental matrix estimation from noisy and inexact keypoint matches is a core computer vision problem. It provides a challenging setting for applying RSR methods. 

We review this setting as follows. Let $(\xbm,\xbm') \in \R^3 \times \R^3$ be a correspondence pair of two points in different images that are projections of the same 3D point in the scene, where $\xbm$ and $\xbm'$ are expressed by homogeneous coordinates of planar points. 
The fundamental matrix $\bF \in \R^{3 \times 3}$ relates these corresponding points and the epipolar lines they lie on as follows:  $\xbm'^\top \bF\xbm = 0$~\cite{hartley2003multiple}, or equivalently, 
\begin{align}\label{eq:epipolar_constraint}
    \vecc(\bF)\cdot \vecc(\xbm\xbm'^\top)=0.
\end{align}
where $\vecc(\cdot)$ denotes the vectorized form of a matrix. 
Therefore, ideally, the set of all vectors in $\R^9$ of the form $\vecc(\xbm\xbm'^\top)$, where $(\xbm,\xbm') \in \R^3 \times \R^3$ is a correspondence pair, lies on an 8-subspace in $\R^9$ and its orthogonal complement yields the fundamental matrix. In practice, the measurements of correspondence pairs can be highly corrupted due to poor matching. Moreover, some choices of correspondence pairs and the corruption mechanism may lead to concentration on low-dimensional subspaces within the desired 8-subspace. Furthermore, the corruption mechanism can lead to nontrivial settings of outliers. 
Lastly, since $d=8$ and $D=9$, the theoretical threshold of \cite{hardt2013algorithms} translates to having the fraction of inliers among all data points be at least $8/9\approx 88.9\%$.

Therefore, this application is often a very challenging setting for direct RSR methods.  The best performing RSR methods to date for fundamental matrix estimation are variants of RANSAC~\cite{fischler1981random}. RANSAC avoids any subspace-modeling assumptions, but estimates the susbspace based on testing myriads of samples, each having 7 or 8 point correspondences~\cite{hartley2003multiple}. 

We test the performance of STE in estimating the fundamental matrix on the Photo Tourism database~\cite{snavely2006photo}, where the image correspondences are obtained by SIFT feature similarities~\cite{lowe2004distinctive}. We compare STE with the following 3 top RSR performers according to~\cite{lerman2018overview}: FMS~\cite{lerman2018fast}, spherical FMS (SFMS)~\cite{lerman2018fast} and  TME~\cite{tyler1987distribution,zhang2016robust}. We also compared with vanilla RANSAC~\cite{fischler1981random} and two of its specialized extensions, which are  
state-of-the-art performers for estimating fundamental matrices: 
locally optimized RANSAC (LO-RANSAC)~\cite{chum2003locally}
and degeneracy-check enabled RANSAC (DEGENSAC)~\cite{chum2005two}. 
For the RSR methods we used codes from the supplementary material of~\cite{lerman2018overview} with their default options. We further used the Python package \texttt{pydegensac} for implementing LO-RANSAC and DEGENSAC with the inlier threshold $\eta=0.75$. 
For STE, we used \Cref{alg:ste-lambda} to estimate the best $\gamma$ from $\{(2i)^{-1}\}_{i=1}^5$.

We measure the accuracy of the results according to the median and mean errors of relative rotation and direction vectors directly obtained by the fundamental matrices for each method. For computing these errors, we compared with ground-truth values
provided by \cite{snavely2006photo,wilson2014robust}. \Cref{fig:fund_plot} describes the result of the mean errors for relative rotation per dataset of Photo Tourism, where the other three errors and  mAA($10^\circ$) are in the supplemental material. STE is significantly better than top RSR performers (TME, FMS and SFMS). Overall, it appears that STE performs better than vanilla RANSAC, except for the Ellis Island and Vienna Cathedral datasets, where RANSAC outperforms STE. 
STE is still competitive when compared with LO-RANSAC and DEGENSAC, except for Notre Dame and the latter two datasets. 

\subsection{Initial Camera Removal for SfM}\label{sec:sfm}

\begin{figure*}[htbp]
    \centering 
    \makebox[\textwidth][c]{\includegraphics[width=1\textwidth]{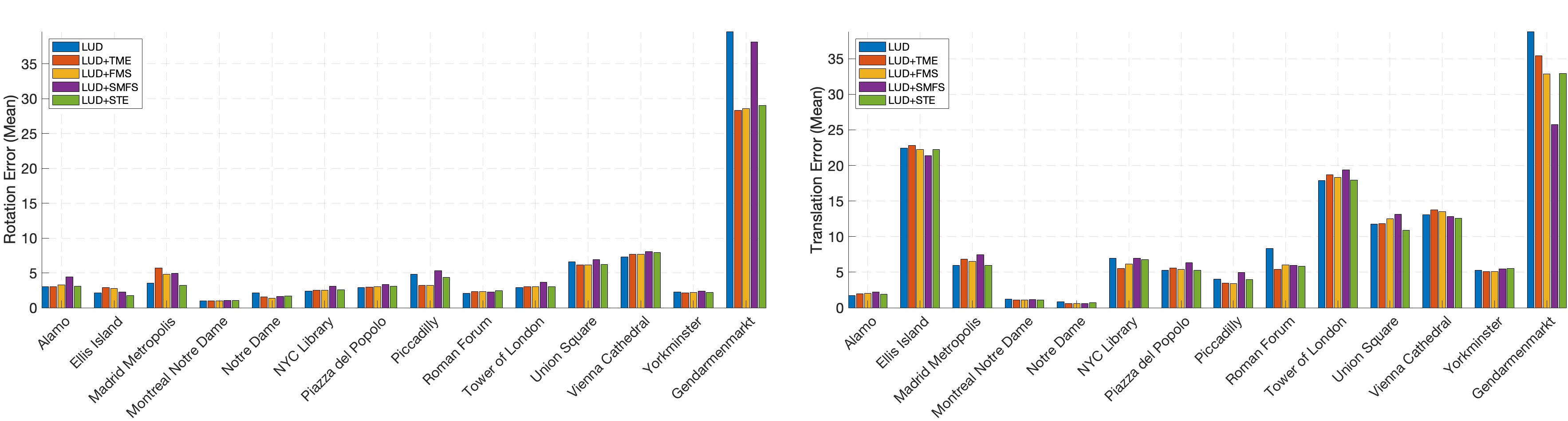}}
    \caption{Mean (absolute) rotation errors (in degrees, left) and mean translation errors (in degrees, right) of LUD and four RSR methods used to initially screen bad cameras within LUD applied to the 14 datasets of Photo Tourism.}
    \label{fig:SfM_screening_median_runtime}
\end{figure*}

We propose a novel application of RSR for SfM and test STE for this application. Even though our framework is not sufficiently practical at this point, it allows testing STE in a different setting where $N=D$ is very large and $d=6$.
Our idea is to use RSR within the SfM pipeline right after estimating the fundamental matrices, in order to remove some cameras that result in inaccurate estimated fundamental matrices. The hope is that eventually such methods may reduce corruption and speed up the costly later computationally intensive stages of the global SfM pipeline. 

There are two main reasons to question such a process. One may first question the gain in improving accuracy. Indeed, since the rest of the pipeline already identifies corrupted pairwise measurements, this process may not improve accuracy and may even harm it as it removes whole cameras and not pairs of cameras. That is, it is possible that a camera, which results in bad pairwise measurement, also contributes to some other accurate pairwise estimates that can improve the overall accuracy.  
The second concern is in terms of speed. In general, the removal of cameras may result in higher or comparable speed. Indeed, the LUD global pipeline \cite{ozyesil2015robust}, which we follow, examines the parallel rigidity of the viewing graph and extracts the maximal parallel rigid subgraph. Thus earlier removal of cameras may worsen the parallel rigidity of the graph and increase the computation due to the need of finding a maximal parallel rigid subgraph. 
For example, \cite{shi21scc} removes cameras in an earlier stage of the LUD pipeline, but results in higher computational cost than the LUD pipeline.
Therefore, improvement of speed for the LUD pipeline by removing cameras is generally non-trivial. Moreover, currently we use scale factors obtained by first running LUD, so we do not get a real speed improvement.
Nevertheless, the proposed method is insightful 
whenever it may indicate clear improvement in accuracy for a dataset, since one can then infer that the current pipeline is not effective enough in handling corrupted measurements, which can be easily recognized by a simple method. Furthermore, improvement in ``speed'' can be indicative of maintaining parallel rigidity.

Our RSR formulation is based on a fundamental observation by Sengupta et al.~\cite{sengupta2017new} on the low-rank of the $n$-view essential (or fundamental) matrix. The $n$-view  essential matrix $\bE$ of size $3n \times 3n$ is formed by stacking all $\binom{n}{2}$ essential matrices, while being appropriately scaled. That is, the $ij$-th block of $\bE$ is the essential matrix  for the $i$-th and $j$-th cameras, where each $\bE_{ij}$ is scaled by a factor $\lambda_{ij}$ in accordance with the global coordinate system (see \cite{sengupta2017new,kasten2019algebraic,kasten2019gpsfm}).  
It was noticed in \cite{sengupta2017new}
that $\bE$ has rank 6. Moreover, \cite{sengupta2017new} characterized the set of $n$-view essential matrices whose camera centers are not all collinear by the satisfaction of a few algebraic conditions, where the major one is $\rank(\bE)=6$. Further explanation appears in \cite{kasten2019algebraic}. 

We propose a straightforward application of RSR, utilizing these ideas to initially eliminate cameras that introduce significant corruption to the essential matrices.  
For this purpose, we compute the essential matrices (by computing first the fundamental matrices and then using the known camera calibration) and scale each matrix according to the factor obtained by the LUD pipline \cite{ozyesil2015robust} (note that this is the initial scaling applied in \cite{sengupta2017new,kasten2019algebraic,kasten2019gpsfm} before applying a non-convex and nontrivial optimization procedure that refines such scales). Using these appropriately scaled essential matrices, we form the $n$-view essential matrix $\bE$ of size $3n \times 3n$.  
We denote the $3n \times 3$ column blocks of $\bE$ by 
$\bE_{:,1}$, $\ldots$,  $\bE_{:,n}$ (since $\bE$ is symmetric they are the same as the row blocks transposed).
We treat $\bE$ as a data matrix with $D=N=3n$, where the columns of $\bE$ are the data points. We apply RSR with $d=6$, recover a $d$-dimensional robust subspace and identify the outlying columns whose distance is largest from this subspace.  To avoid heuristic methods for the cutoff of outliers we assume a fixed percentage of $20\%$ outlying columns.  
If a column block, $\bE_{:,i}$ contains an outlying column, we remove its corresponding camera $i$. Consequently, a smaller percentage of cameras (about $10-15\%$) will be eliminated. 

We use the Photo Tourism database~\cite{snavely2006photo} with precomputed pairwise image correspondences provided by~\cite{sengupta2017new} (they were obtained by thresholding SIFT feature similarities). 
To compute scale factors for the essential matrices we use the output of the LUD pipeline~\cite{ozyesil2015robust}
as follows (following an idea proposed in~\cite{sengupta2017new} for initializing these values):
Given the essential matrix for cameras $i$ and $j$ computed at an early stage of our pipeline, $\bE_{ij}$,  and the one obtained by the full LUD pipeline, $\bE_{ij}^{\text{LUD}}$, the scaling factor is $\lambda_{ij}={\langle \bE_{ij},[\bE_{ij}^{\text{LUD}}]\rangle}/{\|[\bE_{ij}^{\text{LUD}}]\|^2_F}$. Since many values of $\bE_{ij}$ are missing, we also apply matrix completion.   

We compare the LUD pipeline with the LUD pipeline combined with the filtering processes achieved by STE, FMS, SFMS, and TME. For STE we fix $\gamma =1/3$, though any other value we tried yielded the same result. 
We report both mean and median errors of rotations and translations and runtime of the standard LUD and the RSR+LUD methods with initial screening of 
cameras. \Cref{fig:SfM_screening_median_runtime} shows the mean rotation and translation errors, where the rest of the figures and a summarizing table are in the supplementary material.  
In general, STE demonstrates slightly higher accuracy compared to other RSR methods. Improved accuracy is particularly notable when matrix completion is not utilized, as demonstrated in the supplementary material.
We observe that LUD+STE generally improves the estimation of camera parameters (both rotations and translations) over LUD. The improvement of LUD+STE is noticeable in Roman Forum and Gendarmenmarkt. In the supplementary material we show further improvement for Gendarmenmarkt with the removal of $45\%$ outlying columns. While the resulting errors are still large, their improvement shows some potential in dealing with difficult SfM structure by initially removing cameras in a way that may help eliminate  some scene ambiguities, which are prevalent in Gendarmenmarkt.
In terms of runtime, both LUD+STE and LUD+SFMS demonstrate significant improvements, where LUD+SFMS is even faster than LUD+STE. While this does not yet imply faster handling of the datasets (as we use initial scaling factors obtained by LUD), it indicates the efficiency of the removal of outliers in maintaining  parallel rigidity.

\section{Conclusions}
\label{sec:conclusions}
We introduce STE, a meticulously crafted adaptation of TME designed to address challenges within RSR. Theoretical guarantees demonstrate its ability to recover the true underlying subspace reliably, even with a smaller fraction of inliers compared to the well-known theoretical threshold. Under the generalized haystack model, we show that this initialization can be chosen as TME itself, leading to improved handling of a smaller fraction of inliers compared to TME. Our exploration extends to practical applications, where STE proves effective in two 3D vision tasks: robust fundamental matrix estimation and screening of bad cameras for improved SfM.

Several avenues for future research include: $\bullet$ Exploring adaptations of other robust covariance estimation methods to RSR. $\bullet$ Studying effective initialization for STE both in theory and in practice. $\bullet$ In-depth theoretical exploration of the optimal choice of the parameter $\gamma$. $\bullet$ Study of alternative ways of adapting TME to RSR problems. 
$\bullet$ Improving STE for fundamental matrix estimation following ideas similar to those in \cite{chum2005two,frahm2006ransac,raguram2012usac} for addressing challenging degeneracies.
$\bullet$ Enhancing our initial idea of initial removal of bad cameras, specifically attempting to use it to rectify challenging scene ambiguities. 
$\bullet$ 
Testing our methods for SfM using more recent feature matching algorithms.


{
    \small
    \bibliographystyle{ieeenat_fullname}
    \bibliography{RSR-ref}
}


\newpage
\appendix
\onecolumn

{\centering\huge{Supplementary Material}}
\vspace{.2in}

Our theoretical results are proved and further explained in \cite{lerman2024theoretical}.
Additional numerical results and discussion are provided in \S\ref{sec:supp_numerical}.
We refer to references, equation numbers and sections from the main text. We also follow the same numbering of figures, tables and equations, and thus start, for example, with Figure 3.

\section{Supplementary Details for the Numerical Experiments}
\label{sec:supp_numerical}

\begin{figure*}[htbp]
    \centering
    \includegraphics[width=1\linewidth]{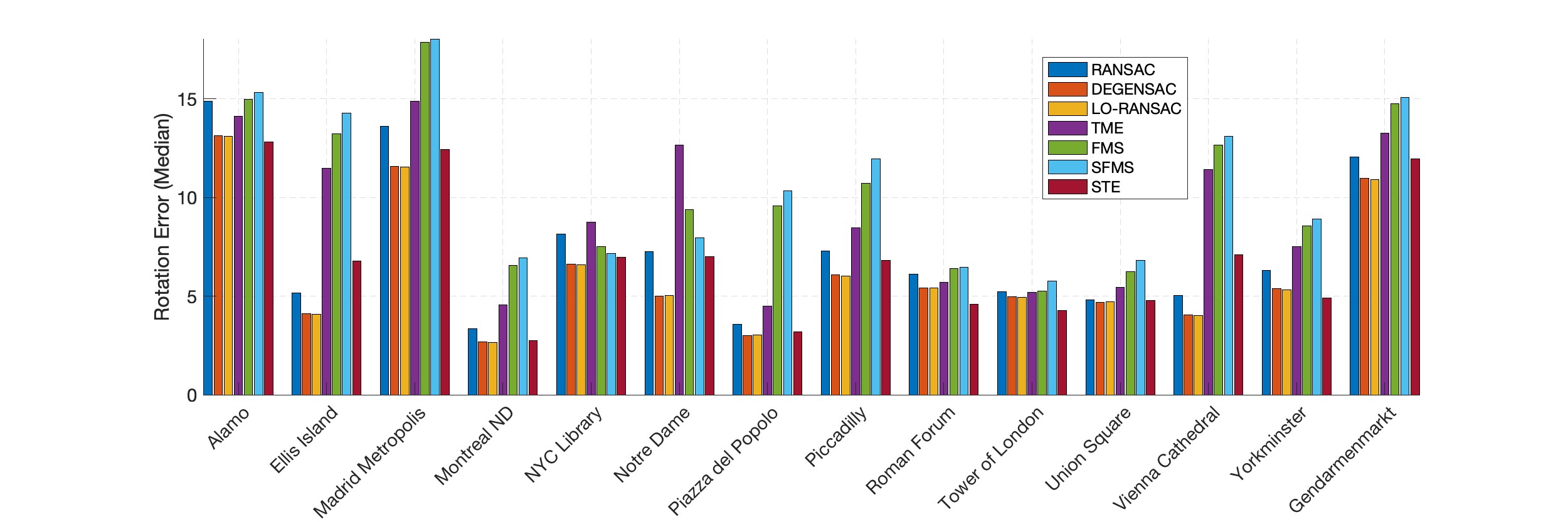}
    \includegraphics[width=1\linewidth]{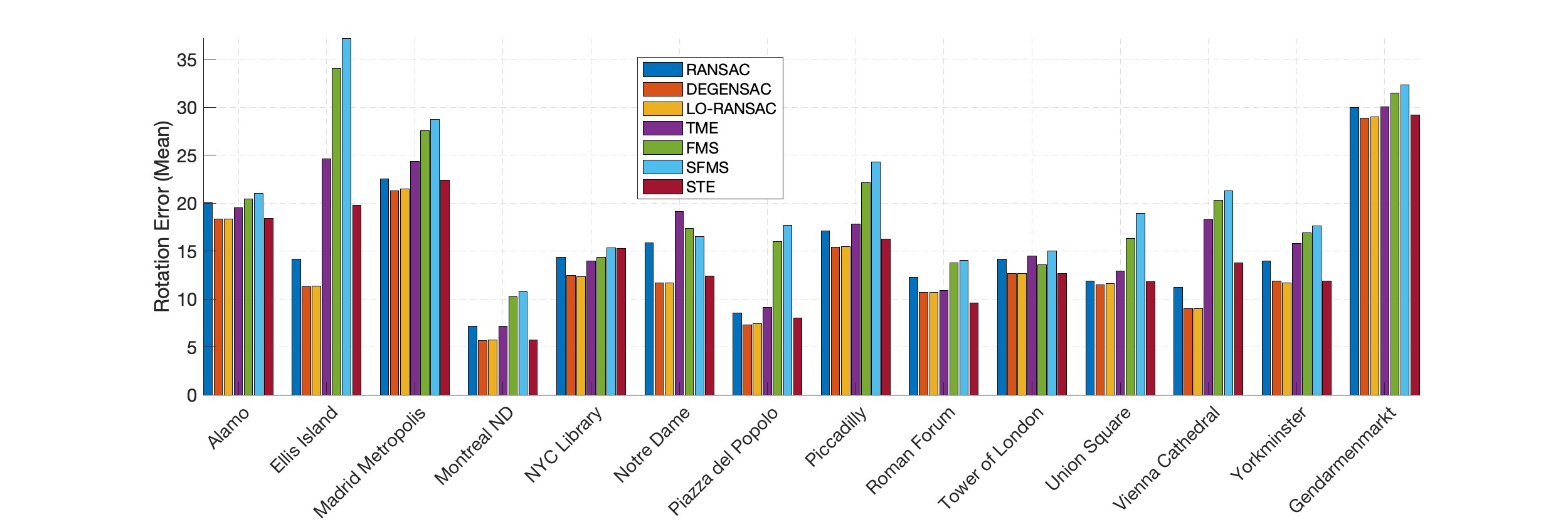}
    \caption{Median (top) and mean (bottom) relative rotation errors  (in degrees)  obtained by seven algorithms for the 14 datasets of Photo Tourism. The numerical results are reported in \Cref{tab:fund_rotation}}
    \label{fig:fund_plot_rotation}
\end{figure*}

\begin{figure*}[htbp]
    \centering
    \includegraphics[width=1\linewidth]{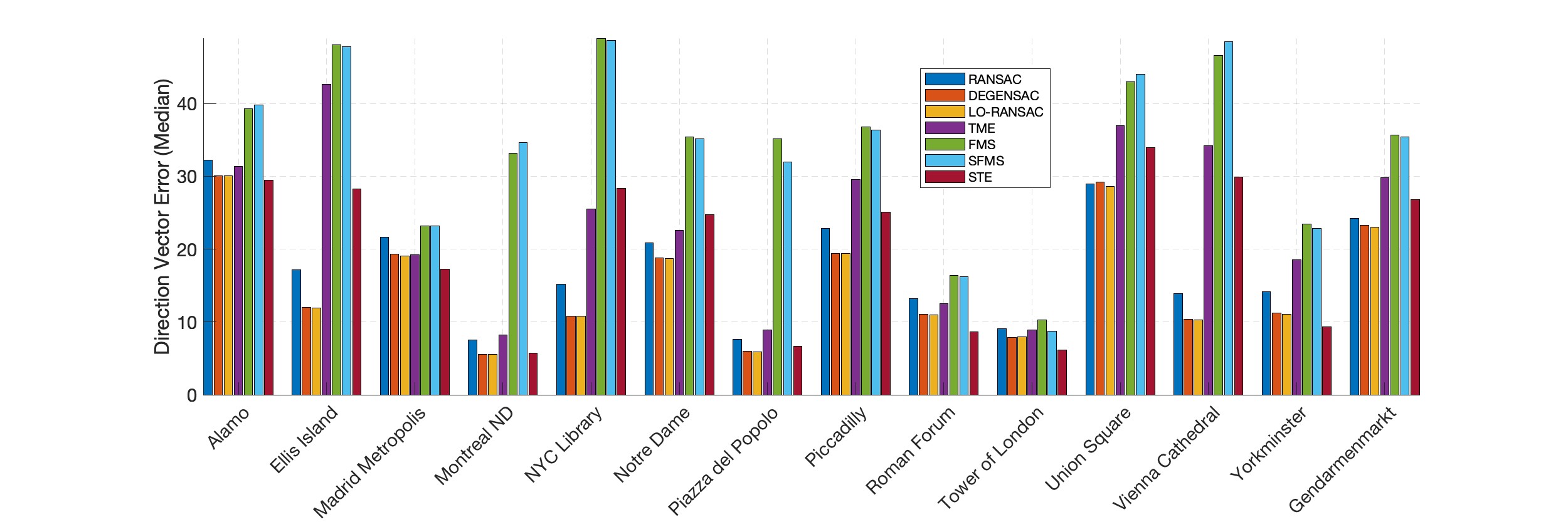}
    \includegraphics[width=1\linewidth]{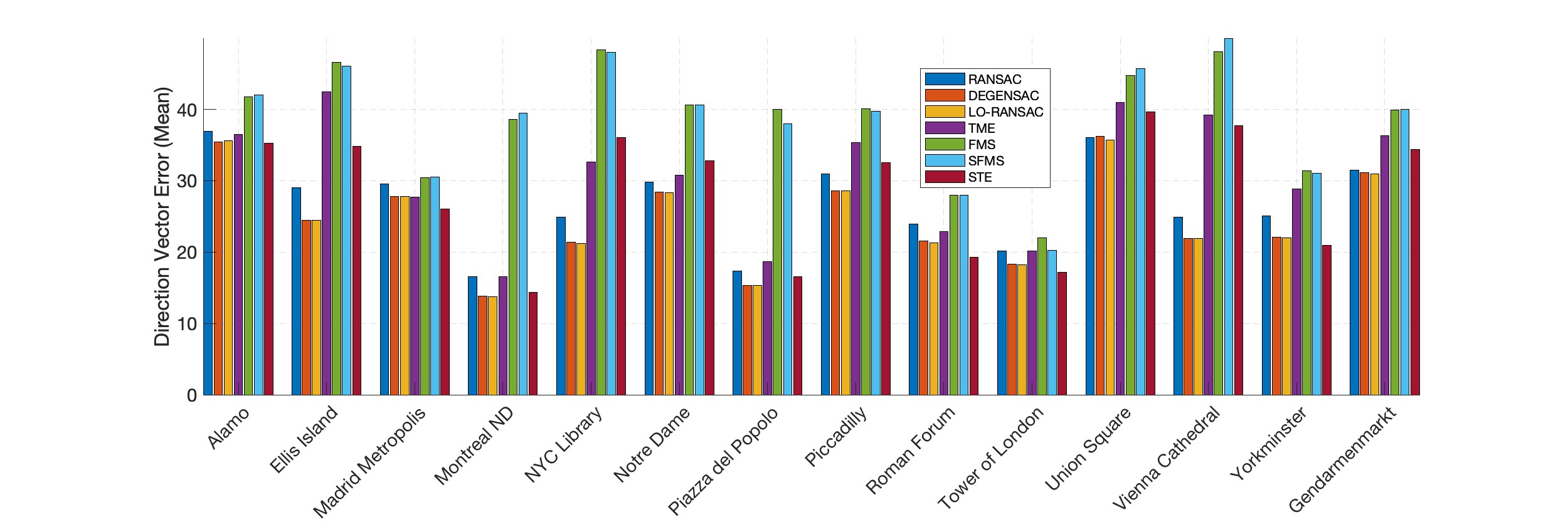}
    \caption{Median (top) and mean (bottom) errors of direction vectors (in degrees) obtained by seven algorithms for the 14 datasets of Photo Tourism. The numerical results are reported in \Cref{tab:fund_translation}}
    \label{fig:fund_plot_translation}
\end{figure*}

We provide additional numerical results and details for our two different types of applications in \S\ref{sec:numerics_1st} and \S\ref{sec:numerics_2nd}.

\subsection{Supplementary Details for Fundamental Matrix Estimation}
\label{sec:numerics_1st}

We review some details about data normalization applied for RSR methods in \S\ref{sec:normalize} and the implementation of the RSR methods in \S\ref{sec:impl_FM}. We  provide additional numerical results in \S\ref{sec:supp_FM_numerics}, and runtime comparison of only the RSR algorithms (including RANSAC) in \S\ref{sec:runtime}. 

\subsubsection{Details of Data Normalization}
\label{sec:normalize}
Fundamental matrix estimation requires proper normalization of the input data~\cite{hartley2003multiple}. We chose to use a normalization that is very similar to the commonly used one in the 8-point algorithm; nevertheless, we apply the normalization to the full dataset as our method is applied to the full dataset. We notice that there are even slightly better normalization techniques, however, to make sure that our competitive performance is due to the proposed method and not the normalization itself, we use the direct generalization of the standard method.

Each data point is represented by a column vector of homogeneous coordinates, $\xbm\in\Rbb^3=(x_1,x_2,1)^\top$. For the $N$ data points, $\xbm_1,\ldots,\xbm_N$, we form  the data matrices $\bX = [\xbm_1,\ldots,\xbm_N]\in\Rbb^{3\times N}$.
    Let $(\mu_1,\sigma_1)$ and $(\mu_2,\sigma_2)$ be the means and standard deviations of the first and the second rows of $\Xbf$, respectively. 
    We normalize the points as follows: $$\hat{\xbm}_i = \Tbf\xbm_i, \ 1\leq i\leq N,$$ where the normalizing  transformation $\Tbf\in\Rbb^{3\times 3}$ is given by
    \begin{align*}
        \Tbf = 
        \begin{bmatrix}
            \sigma_1^{-1} & 0 & 0 \\ 0 & \sigma_2^{-1} & 0 \\ 0 & 0 & 1
        \end{bmatrix}
        \begin{bmatrix}
            1 & 0 & -\mu_1 \\ 0 & 1 & -\mu_2 \\ 0 & 0 & 1
        \end{bmatrix} = 
        \begin{bmatrix}
            \sigma_1^{-1} & 0 & -\sigma_1^{-1}\mu_1 \\ 0 & \sigma_2^{-1} & -\sigma_2^{-1}\mu_2 \\ 0 & 0 & 1
        \end{bmatrix}.
    \end{align*}
This normalization is applied to any RSR method that uses all data points at once to form the subspace. In our experiments, these methods include FMS, SFMS, TME and STE. For RANSAC and its variants 
(which compute subspaces for 8 points at a time) we use the common normalization \cite{hartley2003multiple}, which is exactly the normalization above, but applied to the 8 points chosen each time.

\subsubsection{Review of the Implementation of the RSR  Algorithms}
\label{sec:impl_FM}
We apply each RSR method (TME, FMS, SFMS and STE) as follows. 
Given two normalized data matrices (see Section \ref{sec:normalize})
$\hat{\bX} = [\hat{\xbm}_1,\ldots,\hat{\xbm}_N]\in\Rbb^{3\times N}$ and $\hat{\bX}' = [\hat{\xbm}_1',\ldots,\hat{\xbm}_N']\in\Rbb^{3\times N}$ whose columns are normalized feature points for the two given images, we form the data matrix $\tilde{\Xbf}=[\tilde{\xbm}_1,\ldots,\tilde{\xbm}_N]\in\Rbb^{9\times N}$ with the following columns: $\tilde{\xbm}_i = \vecc(\hat{\xbm}_i\hat{\xbm}_i'^\top)$, $i=1, \ldots, N$.  
We apply each RSR method 
with $d=8$ and $D=9$ (that is, we aim to recover an 8-subspace in $\R^9$) to $\tilde{\Xbf}$. 
We find the orthogonal vector to this 8-dimensional subspace in $\R^9$ and reshape it in into a $3$ by $3$ matrix $\tilde{\bF}$. 
In order to report a proper fundamental matrix with rank 2, we replace its lowest singular value with 0. Finally, we obtain the estimated fundamental matrix for the original data (before normalization) as follows: $\Tbf^\top\hat{\bF}\Tbf'$, where $\Tbf$ and $\Tbf'$ are the normalization transformations defined in \Cref{sec:normalize}.

\subsubsection{Additional Numerical Results for Fundamental Matrix Estimation}
\label{sec:supp_FM_numerics}

We assess the quality of the fundamental matrix estimation by the median and mean errors of relative rotation and direction vectors directly obtained by the fundamental matrices for the various methods. We recall that these methods include STE (our proposed algorithm), TME~\cite{tyler1987distribution,zhang2016robust}, FMS~\cite{lerman2018fast}, SFMS~\cite{lerman2018fast}, vanilla RANSAC~\cite{fischler1981random}, DEGENSAC~\cite{chum2005two} and LO-RANSAC~\cite{chum2003locally}. We first explain how we compute the above errors for any of these methods.

For any two cameras, $i$ and $j$, 
We estimate the fundamental matrix $\widetilde{\bF}_{ij}$ from the correspondence pairs of the images of these cameras and then extract from $\widetilde{\bF}_{ij}$ the relative rotation,  $\widetilde{\Rbf}_{ij}$, and the direction vector (that is, relative translation normalized to have norm 1) in the coordinates of camera $i$, $\widetilde{\tbm}_{ij}$~\cite{hartley2003multiple}. We remark that we need to normalize the relative translation to obtain direction vectors due to the scale ambiguity of the fundamental matrix. 
For cameras $i$ and $j$, the relative rotation error compares  the estimated relative rotation, $\widetilde{\Rbf}_{ij}$, 
and the ground-truth one (provided by the Photo Tourism database), $\{\Rbf_{ij}^*\}$, as follows:
$$e_R = \cos^{-1}
\left(
\frac{\tr
\left({\Rbf_{ij}^*}^\top\widetilde{\Rbf}_{ij}\right)-1}{2}
\right).$$

Note that the estimated direction vectors, $\{\widetilde{\tbm}_{ij}\}$, are defined up to a global orientation and in order to align them with the ground-truth information (given in the frame of camera $i$) obtained from the Photo Tourism data set, $\{\tbm^*_{ij}\}$, we find a rotation matrix $\Rbf_{\ali}$, which minimizes $\sum_{1\leq i,j\leq N}\|\tbm^*_{ij}-\Rbf_{\ali}\widetilde{\tbm}_{ij}\|^2_F$. 
For cameras $i$ and $j$, the error of direction vectors is
$$e_T = \cos^{-1}(|\tbm_{ij}^*\cdot \Rbf_{\ali}\widetilde{\tbm}_{ij}|).$$We remark that the estimated direction vector, $\widetilde{\tbm}_{ij}$ in the frame of camera $i$ implicitly requires estimating both the direction vector in world's coordinates and the absolute rotation matrix (up to global scale) that maps from world's coordinates to the frame of camera $i$, and thus seems to be susceptible to large errors.

\begin{table*}[htbp]
\centering
\resizebox{\linewidth}{!}{%
\begin{tabular}{l|cc|cc|cc|cc|cc|cc|cc}
\toprule
\multirow{2}{*}{Location} & \multicolumn{2}{c|}{RANSAC} & \multicolumn{2}{c|}{DEGENSAC} & \multicolumn{2}{c|}{LO-RANSAC} & \multicolumn{2}{c|}{TME} & \multicolumn{2}{c|}{SFMS} & \multicolumn{2}{c|}{FMS} & \multicolumn{2}{c}{STE} \\
& $\tilde{e}_R$ & $\hat{e}_R$ & $\tilde{e}_R$ & $\hat{e}_R$ & $\tilde{e}_R$ & $\hat{e}_R$ & $\tilde{e}_R$ & $\hat{e}_R$ & $\tilde{e}_R$ & $\hat{e}_R$ & $\tilde{e}_R$ & $\hat{e}_R$ & $\tilde{e}_R$ & $\hat{e}_R$ \\
\midrule
Alamo             & 14.90 & 20.02 & 13.13 & 18.38 & 13.12 & 18.35 & 14.13 & 19.50 & 14.98 & 20.44 & 15.33 & 21.02 & 12.81 & 18.43 \\
Ellis Island      & 5.17  & 14.17 & 4.10  & 11.26 & 4.09  & 11.38 & 11.48 & 24.64 & 13.22 & 34.06 & 14.27 & 37.16 & 6.77  & 19.76 \\
Madrid Metropolis & 13.61 & 22.52 & 11.57 & 21.28 & 11.56 & 21.48 & 14.89 & 24.38 & 17.89 & 27.58 & 18.05 & 28.78 & 12.45 & 22.42 \\
Montreal N.D.     & 3.36  & 7.14  & 2.69  & 5.69  & 2.66  & 5.72  & 4.55  & 7.18  & 6.56  & 10.22 & 6.94  & 10.74 & 2.76  & 5.72  \\
NYC Library       & 8.16  & 14.34 & 6.64  & 12.49 & 6.60  & 12.33 & 8.76  & 13.97 & 7.50  & 14.35 & 7.18  & 15.35 & 6.98  & 15.26 \\
Notre Dame        & 7.25  & 15.84 & 4.99  & 11.67 & 5.03  & 11.68 & 12.67 & 19.16 & 9.40  & 17.36 & 7.96  & 16.54 & 7.02  & 12.42 \\
Piazza del Popolo & 3.59  & 8.54  & 3.00  & 7.32  & 3.02  & 7.45  & 4.51  & 9.10  & 9.59  & 15.98 & 10.35 & 17.69 & 3.20  & 8.02  \\
Piccadilly        & 7.30  & 17.11 & 6.08  & 15.42 & 6.03  & 15.47 & 8.47  & 17.85 & 10.73 & 22.12 & 11.95 & 24.29 & 6.81  & 16.29 \\
Roman Forum       & 6.11  & 12.29 & 5.42  & 10.69 & 5.41  & 10.68 & 5.69  & 10.90 & 6.41  & 13.75 & 6.46  & 14.05 & 4.59  & 9.57  \\
Tower of London   & 5.24  & 14.16 & 4.96  & 12.69 & 4.94  & 12.64 & 5.20  & 14.47 & 5.27  & 13.55 & 5.77  & 15.02 & 4.28  & 12.66 \\
Union Square      & 4.82  & 11.89 & 4.70  & 11.46 & 4.73  & 11.60 & 5.44  & 12.95 & 6.24  & 16.34 & 6.82  & 18.91 & 4.79  & 11.83 \\
Vienna Cathedral  & 5.04  & 11.23 & 4.05  & 8.98  & 4.03  & 9.01  & 11.44 & 18.29 & 12.67 & 20.31 & 13.12 & 21.27 & 7.11  & 13.79 \\
Yorkminster       & 6.31  & 13.97 & 5.39  & 11.85 & 5.33  & 11.70 & 7.52  & 15.83 & 8.55  & 16.93 & 8.92  & 17.63 & 4.92  & 11.89 \\
Gendarmenmarkt    & 12.05 & 29.99 & 10.99 & 28.89 & 10.93 & 28.99 & 13.28 & 30.05 & 14.77 & 31.48 & 15.09 & 32.38 & 11.97 & 29.18
\\
\bottomrule
\end{tabular}
}
\caption{Median and mean relative rotation errors (in degrees) obtained by seven algorithms for the 14 datasets of Poto Tourism. $\tilde{e}_R$ is the median relative rotation error, $\Hat{e}_R$ is the mean relative rotation error.}
\label{tab:fund_rotation}
\end{table*}

\begin{table*}[htbp]
\centering
\resizebox{\linewidth}{!}{%
\begin{tabular}{l|cc|cc|cc|cc|cc|cc|cc}
\toprule
\multirow{2}{*}{Location} & \multicolumn{2}{c|}{RANSAC} & \multicolumn{2}{c|}{DEGENSAC} & \multicolumn{2}{c|}{LO-RANSAC} & \multicolumn{2}{c|}{TME} & \multicolumn{2}{c|}{SFMS} & \multicolumn{2}{c|}{FMS} & \multicolumn{2}{c}{STE} \\
& $\tilde{e}_T$ & $\hat{e}_T$ & $\tilde{e}_T$ & $\hat{e}_T$ & $\tilde{e}_T$ & $\hat{e}_T$ & $\tilde{e}_T$ & $\hat{e}_T$ & $\tilde{e}_T$ & $\hat{e}_T$ & $\tilde{e}_T$ & $\hat{e}_T$ & $\tilde{e}_T$ & $\hat{e}_T$ \\
\midrule
Alamo             & 32.20 & 36.97 & 30.09 & 35.46 & 30.11 & 35.57 & 31.36 & 36.49 & 39.31 & 41.74 & 39.85 & 41.98 & 29.51 & 35.25 \\
Ellis Island      & 17.16 & 29.04 & 11.98 & 24.43 & 11.90 & 24.45 & 42.62 & 42.42 & 48.11 & 46.62 & 47.79 & 46.02 & 28.30 & 34.84 \\
Madrid Metropolis & 21.67 & 29.56 & 19.36 & 27.83 & 19.09 & 27.82 & 19.23 & 27.71 & 23.22 & 30.47 & 23.18 & 30.52 & 17.23 & 26.06 \\
Montreal N.D.     & 7.49  & 16.51 & 5.57  & 13.80 & 5.53  & 13.75 & 8.25  & 16.55 & 33.20 & 38.61 & 34.65 & 39.51 & 5.75  & 14.34 \\
NYC Library       & 15.17 & 24.87 & 10.77 & 21.35 & 10.80 & 21.25 & 25.48 & 32.62 & 48.89 & 48.34 & 48.69 & 48.02 & 28.33 & 36.01 \\
Notre Dame        & 20.91 & 29.83 & 18.82 & 28.40 & 18.75 & 28.30 & 22.61 & 30.74 & 35.38 & 40.63 & 35.16 & 40.60 & 24.77 & 32.83 \\
Piazza del Popolo & 7.58  & 17.35 & 5.95  & 15.31 & 5.93  & 15.34 & 8.88  & 18.70 & 35.17 & 39.99 & 31.98 & 37.96 & 6.70  & 16.57 \\
Piccadilly        & 22.89 & 30.99 & 19.40 & 28.58 & 19.39 & 28.58 & 29.58 & 35.36 & 36.83 & 40.06 & 36.33 & 39.76 & 25.07 & 32.51 \\
Roman Forum       & 13.24 & 23.96 & 11.07 & 21.52 & 10.97 & 21.30 & 12.51 & 22.86 & 16.42 & 27.96 & 16.24 & 27.95 & 8.65  & 19.29 \\
Tower of London   & 9.10  & 20.18 & 7.90  & 18.30 & 7.95  & 18.20 & 8.88  & 20.15 & 10.27 & 22.00 & 8.77  & 20.21 & 6.14  & 17.13 \\
Union Square      & 28.97 & 36.03 & 29.23 & 36.21 & 28.60 & 35.73 & 36.98 & 40.98 & 42.97 & 44.76 & 44.02 & 45.71 & 33.95 & 39.63 \\
Vienna Cathedral  & 13.88 & 24.94 & 10.33 & 21.89 & 10.26 & 21.89 & 34.24 & 39.24 & 46.63 & 48.11 & 48.46 & 49.88 & 29.94 & 37.74 \\
Yorkminster       & 14.17 & 25.11 & 11.20 & 22.05 & 11.05 & 21.99 & 18.54 & 28.87 & 23.48 & 31.42 & 22.82 & 31.03 & 9.32  & 20.98 \\
Gendarmenmarkt    & 24.25 & 31.51 & 23.32 & 31.11 & 23.03 & 30.96 & 29.82 & 36.33 & 35.70 & 39.88 & 35.44 & 40.00 & 26.80 & 34.36 \\
\bottomrule
\end{tabular}
}
\caption{Median and mean errors of direction vectors (in degrees) obtained by seven algorithms for the 14 datasets of Photo Tourism. $\tilde{e}_T$ is the corresponding median error, $\Hat{e}_T$ is the corresponding mean error.}
\label{tab:fund_translation}
\end{table*}

\begin{figure*}[htbp]
    \centering
    \includegraphics[width=0.9\textwidth]{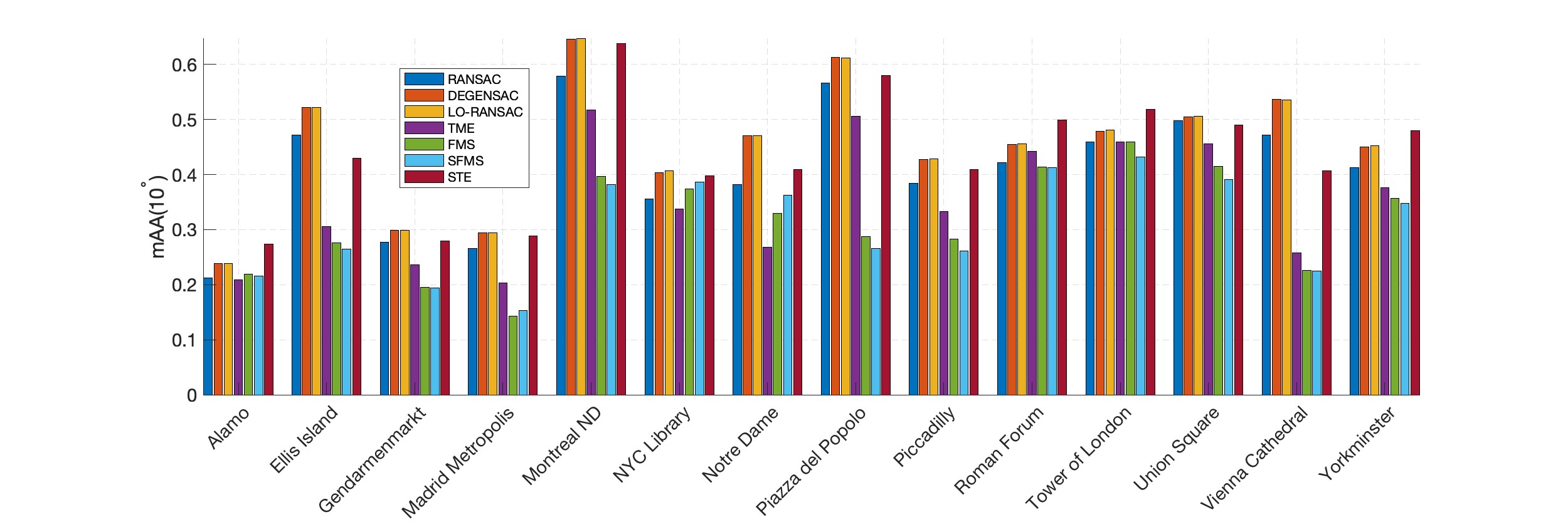}
    \caption{mAA($10^\circ$) obtained by seven algorithms for the 14 datasets of Photo Tourism.}
    \label{fig:mAA_1}
\end{figure*}

\Cref{fig:fund_plot_rotation} is a bar plot presenting the errors of relative rotations (in angles) for the different methods and all 14 datasets
and \Cref{fig:fund_plot_translation}
showcases the errors of direction vectors. 
For completeness, we record the presented numerical values in 
\Cref{tab:fund_rotation} (for relative rotations) and \Cref{tab:fund_translation} (for direction vectors). 
\Cref{fig:mAA_1} presents the mean Average Accuracy (mAA)~\cite{jin2021image} for rotation with threshold $10^\circ$, where as opposed to the former errors which need to be small, higher mAA is better. 

We note that STE significantly outperforms FMS and SFMS. STE also outperforms TME, where the only cases where TME outperforms STE is for direction vector estimation for the NYC Library and Notre Dame datasets. In terms of relative rotation estimation, STE outperforms vanilla RANSAC, where Ellis Island and Vienna Cathedral are the only two datasets where RANSAC outperforms STE. In terms of direction vector estimation, STE outperforms vanilla RANSAC in 7 out of the 14 datasets and in the rest of them vanilla RANSAC outperforms STE. Anyway, the errors of direction vectors are higher than relative rotation errors because of an issue discussed above and many references do not present these errors.   
As shown in \Cref{fig:mAA_1}, STE has a higher mAA($10^\circ$) than other RSR methods. Overall, STE is slightly better than RANSAC. Indeed, the averaged mAA($10^\circ$) for STE is 0.44 for STE and 0.41 for RANSAC. However, RANSAC has a noticeable higher mAA ($10^\circ$) for the following 3 datasets: Ellis Island, Notre Dame and Vienna Cathedral. On the other hand, DEGENSAC and LO-RANSAC are overall slightly better than STE 
(with averaged mAA($10^\circ$) 0.45 for both methods), but STE has a noticeably higher 
mAA ($10^\circ$) for the following 5 datasets:
Alamo, Montreal Notre Dame, Roman Forum, Tower of London and York Minster.

\subsubsection{Runtime Comparison of the RSR components}
\label{sec:runtime}
We compare the elapsed time of the different RSR algorithms, including RANSAC and its variants. Since our current implementation of STE uses Algorithm \ref{alg:ste-lambda} with $m=5$ values of 
$\gamma$ ($\gamma \in \{(2i)^{-1}\}_{i=1}^5$), it is slower than the basic implementation of Algorithm \ref{alg:ste} by a factor 5 . This choice keeps the similar accuracy of STE with more values of $\gamma$ but reduces the overall runtime.

We compared the runtime of STE (with $m=5$) with RANSAC, TME, FMS and SFMS for the 14 Photo Tourism datasets (using Matlab implementations). These experiments were carried out on an Apple Silicon M2. For each dataset, the average runtime among all image pairs is reported in \Cref{tab:runtime_fund}. Overall, the runtimes of STE and RANSAC are somewhat comparable, where the average over all datasets of the runtime of RANSAC is 22.91 msec and of STE is 21.91 msec. However, the variance of the runtime of RANSAC is rather large, unlike STE. 
We further note that TME was the fastest method, however running STE with $m=1$, instead of $m=5$, is slightly faster than TME. 
FMS and SFMS are slower than TME and overall faster than STE (with $m=5$) and RANSAC. 

\begin{table*}[htbp]
\centering
\begin{tabular}{l|ccccc}
\toprule
\multirow{2}{*}{Locations} & \multicolumn{5}{c}{Algorithm Runtimes (ms)} \\ 
\cmidrule{2-6}
& STE & TME & SFMS & FMS & RANSAC \\ 
\midrule
Alamo & 19.2 & 4.0 & 13.7 & 13.2 & 11.8 \\ 
Ellis Island & 19.8 & 4.8 & 9.8 & 9.3 & 44.6 \\ 
Madrid Metropolis & 18.6 & 3.9 & 9.3 & 8.8 & 11.3 \\ 
Montreal N.D. & 18.0 & 3.6 & 18.5 & 17.8 & 31.6 \\ 
NYC Library & 21.1 & 4.5 & 22.5 & 22.0 & 51.7 \\ 
Notre Dame & 19.4 & 4.7 & 18.5 & 17.9 & 15.5 \\ 
Piazza del Popolo & 19.1 & 4.2 & 9.3 & 8.8 & 22.7 \\ 
Piccadilly & 19.0 & 4.4 & 8.8 & 8.2 & 24.0 \\ 
Roman Forum & 18.1 & 3.7 & 16.0 & 15.5 & 7.3 \\ 
Tower of London & 17.6 & 3.7 & 15.4 & 14.8 & 4.5 \\ 
Union Square & 19.0 & 4.2 & 9.6 & 9.1 & 10.2 \\ 
Vienna Cathedral & 20.5 & 4.5 & 21.3 & 20.8 & 37.5 \\ 
Yorkminster & 17.4 & 3.6 & 15.1 & 14.5 & 18.2 \\ 
Gendarmenmarkt & 18.7 & 3.7 & 8.5 & 8.0 & 29.9 \\
\bottomrule
\end{tabular}
\caption{Elapsed runtimes (in milliseconds) of STE, TME, SFMS, FMS and RANSAC for the 14 datasets of Photo Tourism. The runtimes are averaged over all image pairs in the dataset.}
\label{tab:runtime_fund}
\end{table*}

The runtime of RANSAC mainly depends on the fraction of the outliers. For higher fractions of oultiers, RANSAC significantly slowed down compared to other methods, showcasing exponential dependence of time on the fraction of outliers. Indeed, if $\varepsilon$ denotes the fraction of inliers and $d$ the subspace dimension and one requires $99\%$ confidence, the expected number of iterations of RANSAC is given as $\log(1-.99)/\log(1-\varepsilon^d)$~\cite{tordoff2002guided}. 

To further investigate the expected correlation of the runtime and the fraction of outliers, we also considered artificial instances, which we repeatedly generated 100 times.
We fixed $N=400$ and generated two data matrices
${\bX} = [{\xbm}_1,\ldots,{\xbm}_N]\in\Rbb^{3\times N}$ and ${\bX}' = [{\xbm}_1',\ldots,{\xbm}_N']\in\Rbb^{3\times N}$, whose columns contain inlier and outlier feature correspondence pairs in homogenous coordinates as follows. We varied the fraction of designated outliers, where a fixed fraction of columns of ${\bX}$ and ${\bX}'$ (with the same index) were designated as outliers and the rest as inliers. 
For generating an outlier correspondence pair, the first two coordinates of $\xbm_i$ and $\xbm_i'$ were uniformly sampled in $[0,1000]$ and the third coordinates was set to be $1$. 
For an inlier correspondence pair, we followed the procedure of \cite{karimian2023essential} to sample $(\xbm_i,\xbm_i')$ satisfying the epipolar constraint.  

Given the generated matrices $\bX$ and ${\bX}'$, we first normalized them according to the procedure described in \S\ref{sec:normalize} and then used the procedure described in \S\ref{sec:impl_FM} to
estimate the fundamental matrix from these two matrices using any of the RSR methods with $d=8$ and $D=9$. The application of RANSAC is also similar to before. 
The relative rotation was extracted from the estimated fundamental matrix and its error was computed based on the given ground truth rotation for the inliers, which are clarified when following the details of generating them in \cite{karimian2023essential}. Using these 100 samples we computed mAA($10^\circ$).

\begin{figure*}[htbp]
    \centering
    \includegraphics[width=0.45\textwidth]{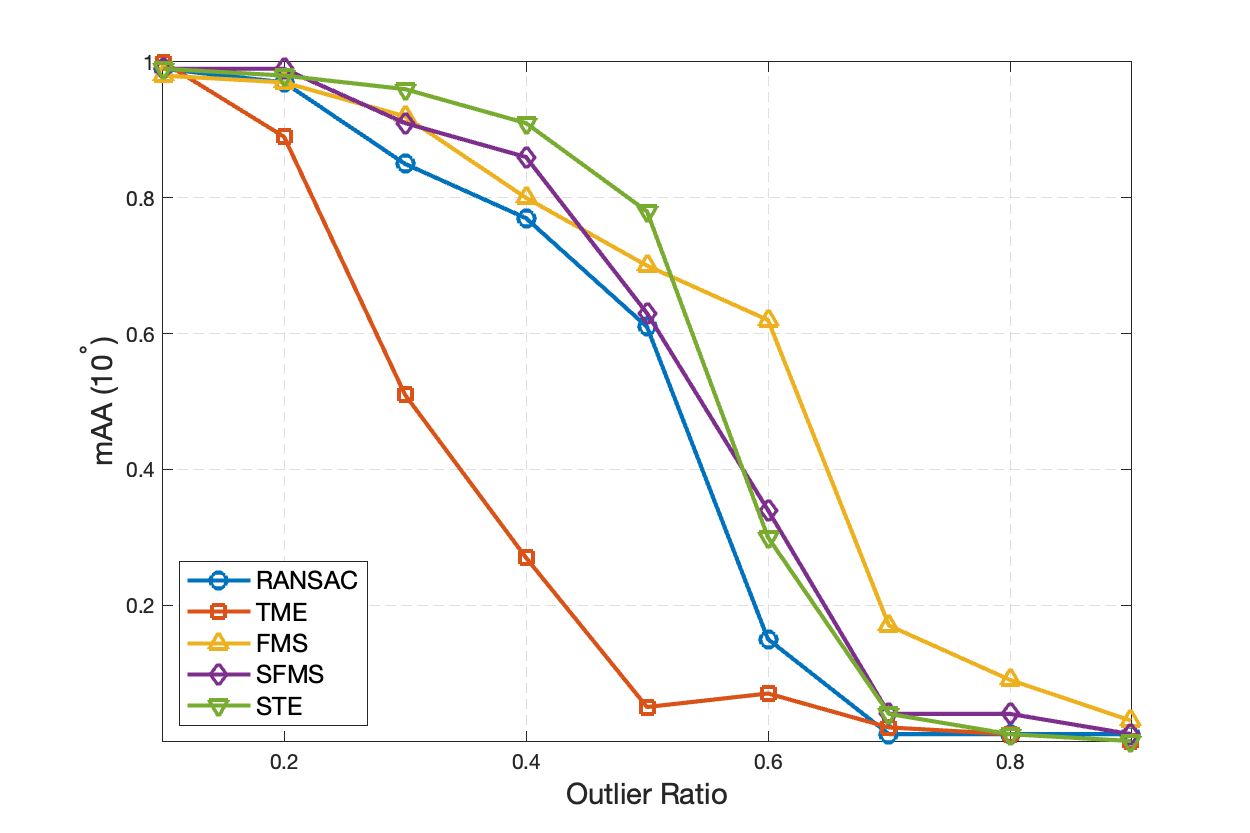}
    \includegraphics[width=0.45\textwidth]{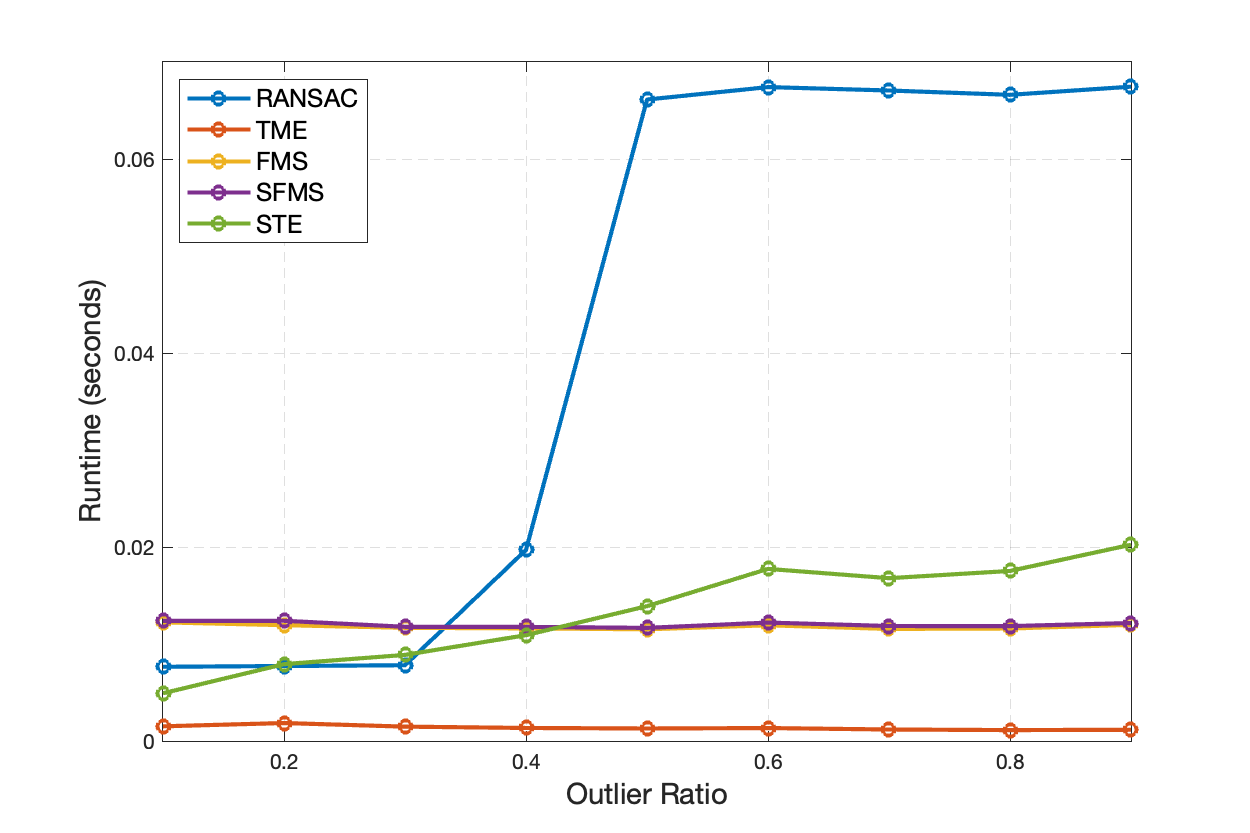}
    \caption{mAA($10^\circ$) (left) and runtime (right, in seconds) of five algorithms for the synthetic feature points data. The $x$-axis indicates the fraction of the outliers, where the total number of points is fixed as $400$. }
    \label{fig:syn_fund}
\end{figure*}

\Cref{fig:syn_fund} reports mAA($10^\circ$) and runtimes of STE (with $m=5$), TME, SFMS, FMS and RANSAC for different fractions of outliers.  The maximum number of iteration of RANSAC was set to be 1000. 
We note that STE (with $m=5$) achieved the highest mAA($10^\circ$) when the fraction of outliers was at most  
$50\%$ and in this case the mAA($10^\circ$) values of FMS are slightly below those of STE. On the other hand, when the fraction of outliers is at least $60\%$ the mAA($10^\circ$) values of STE are higher than STE, but they indicate low accuracy. The values of RANSAC, SFMS and TME are below the ones of STE. Moreover, the mAA($10^\circ$) values of TME are rather low and indicate low accuracy.

Regarding runtime, we first note that FMS, SFMS and TME have fixed times, since the number of data points, $N$, is fixed. Moreover, TME is faster then FMS and SFMS whose time is comparable. On the other hand, STE has a smaller run time for low fraction of outliers, since it converges faster then. Furthermore, the runtime of STE stabilizes when the fraction of outliers is large and it is slightly above FMS and SFMS.
RANSAC is relatively fast for a small fraction of outliers, but it is much slower than all other method when the fraction of outliers is at least $50\%$.

\begin{figure*}[htbp]
    \centering
    \includegraphics[width=0.45\textwidth]{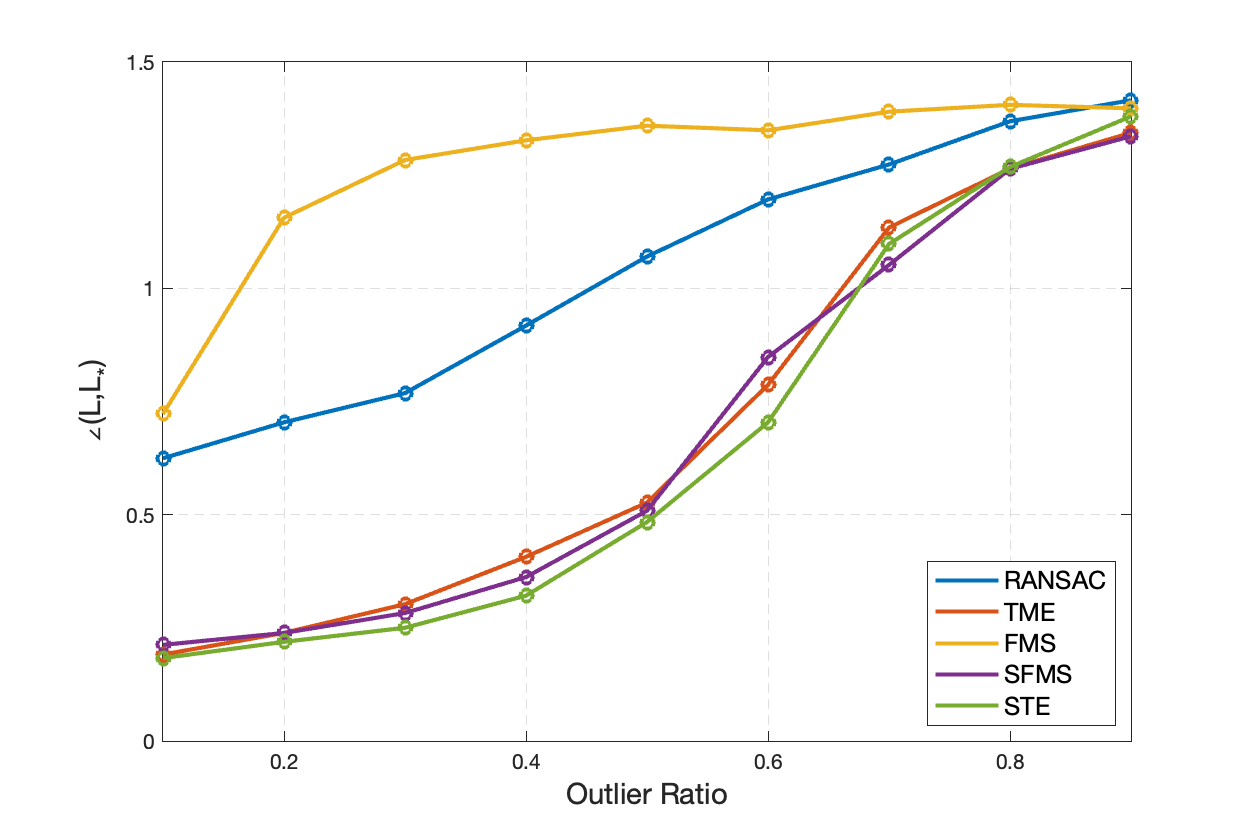}
    \includegraphics[width=0.45\textwidth]{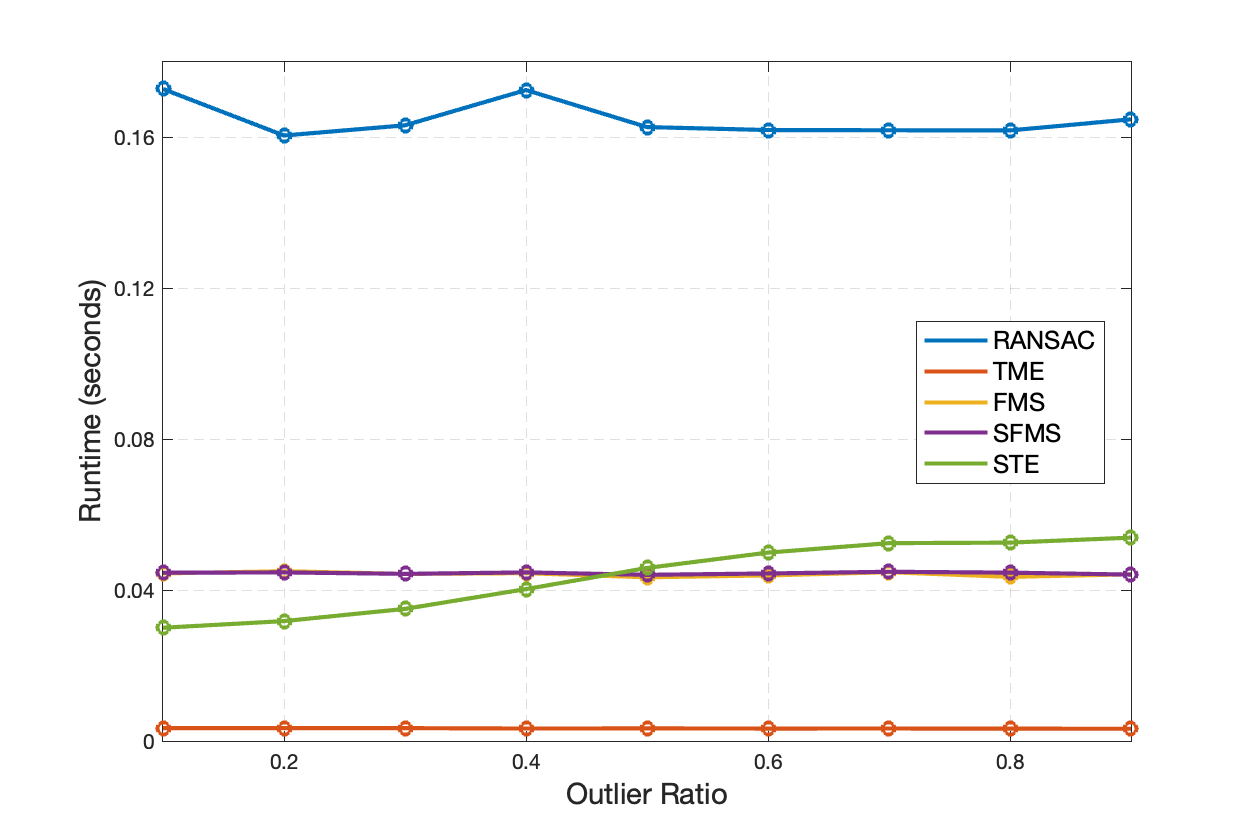}
    \caption{Angular error (left, in radius) and runtime (right, in seconds) of five algorithms for the generalized haystack synthetic data. The error is angle between $L_*$ and the estimated subspace $\hat{L}$ recovered by the algorithm. The $x$-axis is the fraction of outliers.}
    \label{fig:haystack}
\end{figure*}

The above mentioned formula of expected number of iterations implies that RANSAC can easily fail with large $d$. In the setting of 3-view homography estimation, $D=27$ and $d=26$. We thus tried the standard haystack model with $D=27$, $d=26$, $N= 400$ and varied fractions of outliers. 
To measure error, we use the angle between $L_*$, the inlier subspace of the haystack model, and $L$, the estimated subspace. 
\Cref{fig:haystack} compares the errors and runtimes of the five algorithms for  different fractions of outliers. 
We note that STE has the lowest error, but they notably increase for fraction of outliers larger than $50\%$. The errors of SFMS and TME are only slightly above than those of STE. RANSAC performed poorly, even for very small fractions of outliers and FMS is even worse than RANSAC. 

We note that the runtimes of the RSR methods behave similarly to the ones in   \Cref{fig:syn_fund}. In each one of the experiments RANSAC needed the maximal number of iterations, which was set as 1000, and obtained the same runtime, which is significantly larger than the other runtimes.

\subsection{Additional Numerical Results for Initial Camera Removal for SfM}
\label{sec:numerics_2nd}

\begin{figure*}[htbp]
    \centering
    \includegraphics[width=0.9\textwidth]{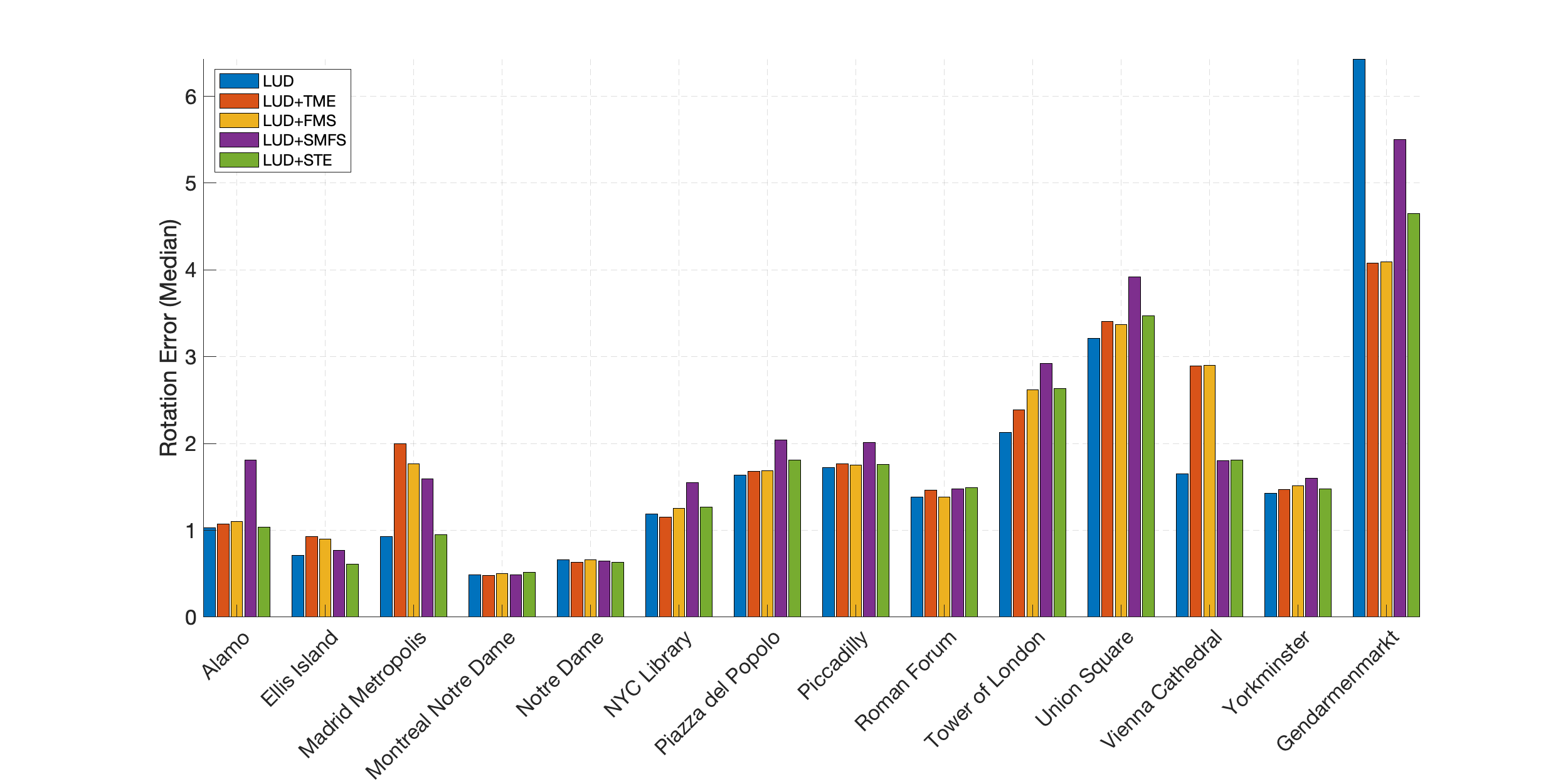}
    \includegraphics[width=0.9\textwidth]{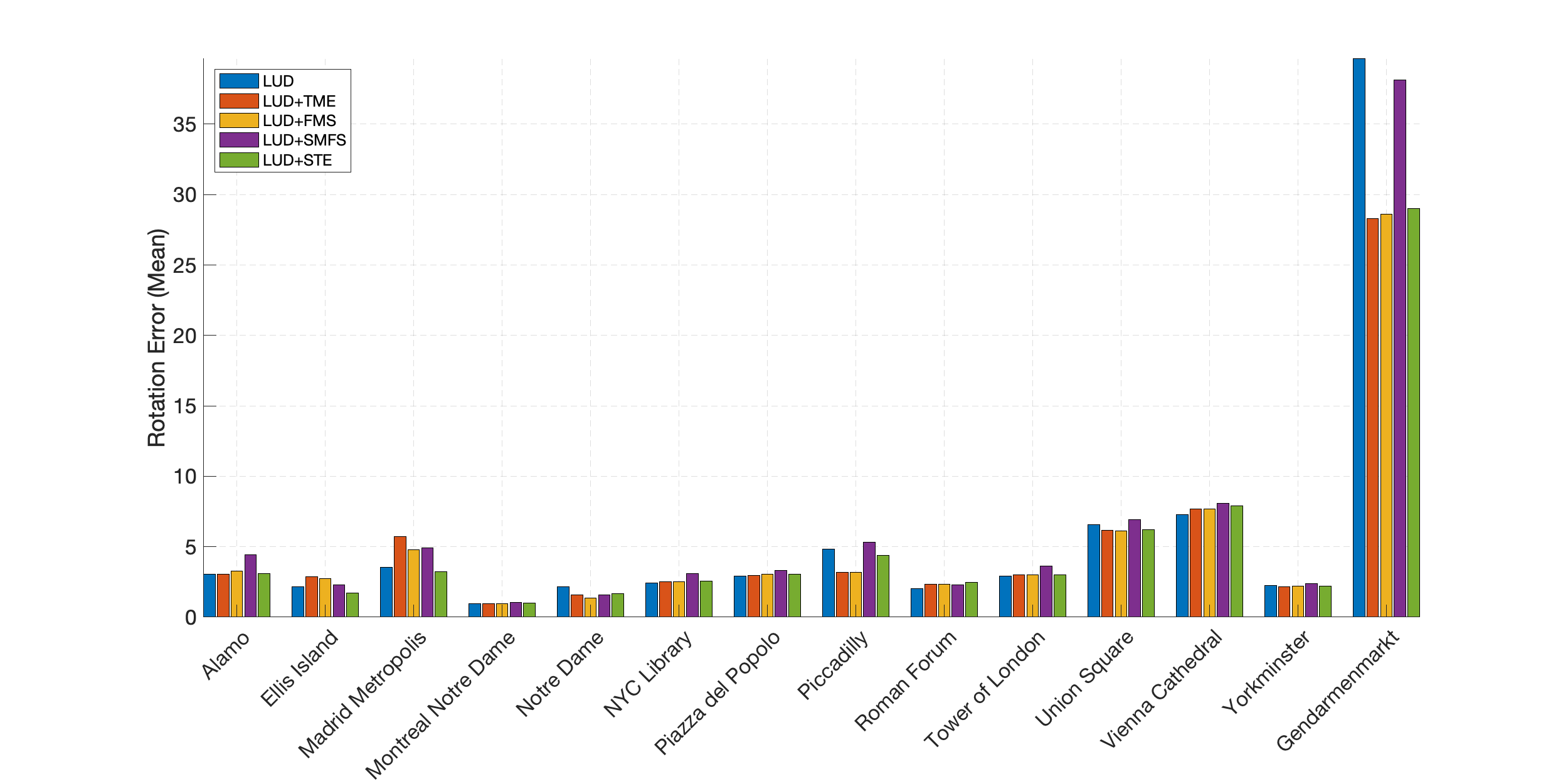}
    \caption{Median (top) and mean (bottom) absolute rotation errors (in degrees) of LUD and four RSR methods used to initially screen bad cameras within LUD applied to the 14 datasets of Photo Tourism. For Gendarmenmarkt, the median rotation error of SFMS was 31.04 degrees and mean error of SFMS was 62.89, which are above the figure range.}
    \label{fig:remove_camera_1}
\end{figure*}

We use all 14 datasets of the Photo Tourism database. As mentioned in the main text, scale factors are first obtained by the LUD algorithm. Many of the blocks $\bE_{ij}$ are absent when there are no matching features between images $i$ and $j$. On the other hand, the geometric relationship between cameras $i$ and $j$ still exists and the ground-truth essential matrix is thus well defined. In such scenarios, a common strategy is to assign to $\bE_{ij}$ a zero matrix~\cite{sengupta2017new}. 
For completeness, we apply our proposed procedure with this strategy in Section \ref{sec:sfm_no_mc}. 
However, the results should be more accurate when using matrix completion to fill the missing blocks of $\bE$. We follow such a strategy here and provide its details as follows. 
Let $\Omega$ denote the set of indices, $ij$, of the blocks $\bE_{ij}$ that can be directly obtained by the images. 
We denote by $\bM\in\Rbb^{3n\times 3n}$ the  desired matrix with blocks $\{\lambda_{ij}\bE_{ij}\mid (i,j)\in\Omega\}$ and additional completed ones. We approximate $\bM$ by solving the following relaxed optimization problem (see e.g., \cite{cai2010singular}): 
\begin{align}\label{eq:Matrix_Completion}
    \widehat{\bM}=\argmin_{\bM\in\Rbb^{3n\times 3n}}\|\bM\|_*,\quad\text{subject to } \bM_{ij}=\lambda_{ij}\bE_{ij},\quad (i,j)\in\Omega.
\end{align}
The Singular Value Thresholding (SVT) algorithm~\cite{cai2010singular} is applied to solve \Cref{eq:Matrix_Completion}. It is worth noting that for some datasets, the sampling ratio $|\Omega|/n^2$ is smaller than $10\%$. To prevent the SVT algorithm from diverging, we set a step size $\delta=n^2/(10|\Omega|)$, which is smaller than the suggested value in \cite[Section 5.1.2]{cai2010singular}. 
Since the block ${\bM}_{ij}$ is defined as an essential matrix, we further project it on the essential matrix manifold. Specifically, we define $\widetilde{\bE}\in\Rbb^{3n\times 3n}$ as follows:
\begin{align*}
    \widetilde{\bE}_{ij} = 
    \begin{cases}
        \lambda_{ij}\bE_{ij}, & \mbox{if }(i,j)\in\Omega \\
        \bP_{E}(\widehat{\bM}_{ij}), & \mbox{if }(i,j)\not\in\Omega \\
        \bm{0} & \mbox{if } i=j,
    \end{cases}
\end{align*}
where for $\widehat{\bM}_{ij}=\bU\Sbf\bV^\top$, $\bP_{E}(\widehat{\bM}_{ij})=\bU\diag([1,1,0])\bV^\top$ is the projector to the essential manifold.  Following the procedures described in \Cref{sec:sfm}, we apply RSR with $d=6$ by treating $\widetilde{\bE}$ as a data matrix of $D=N=3n$, recover a $d$-dimensional robust subspace and identify the outlying columns whose distance is largest from this subspace.

\begin{figure*}[htbp]
    \centering
    \includegraphics[width=0.9\textwidth]{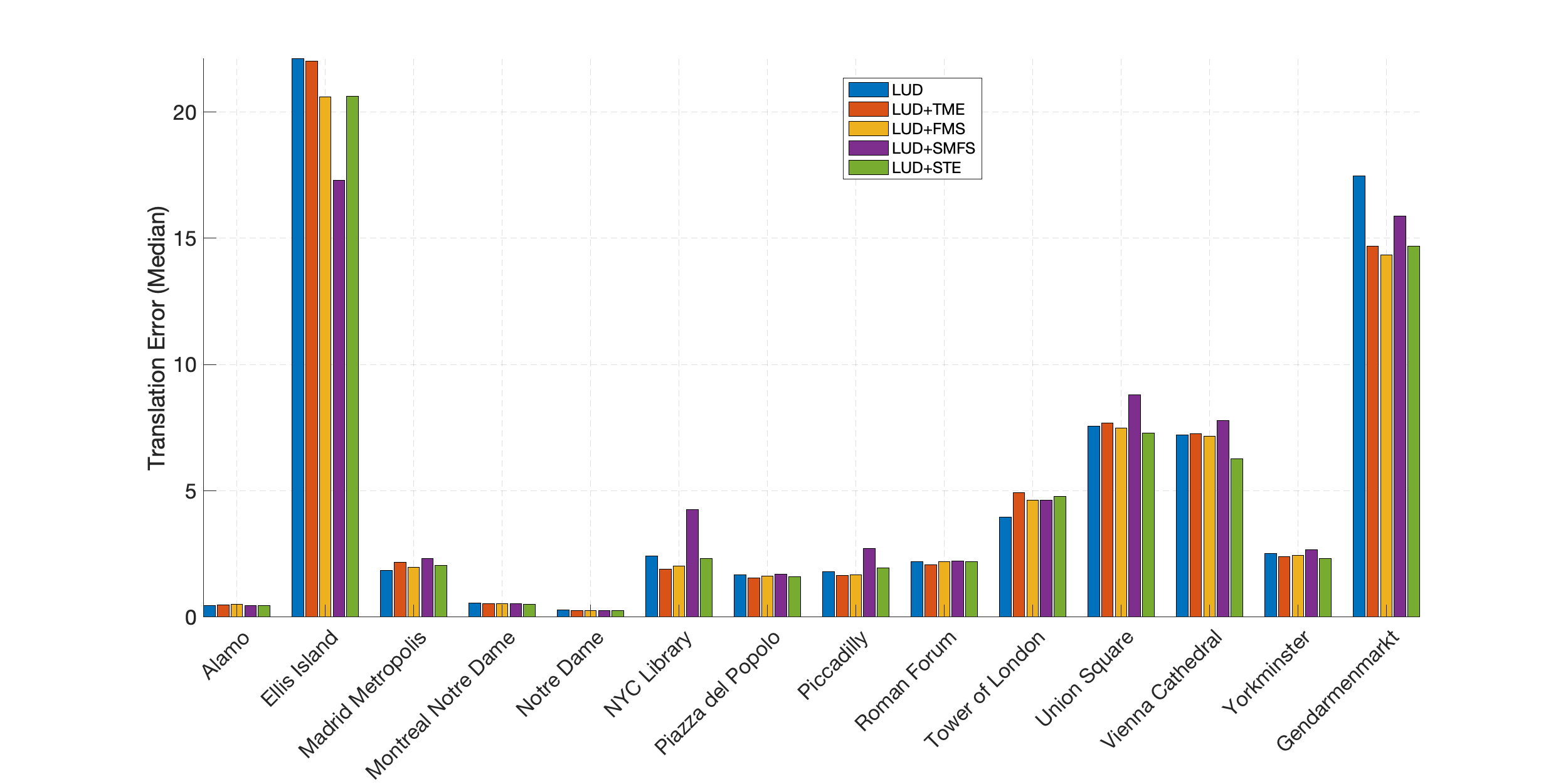}
    \includegraphics[width=0.9\textwidth]{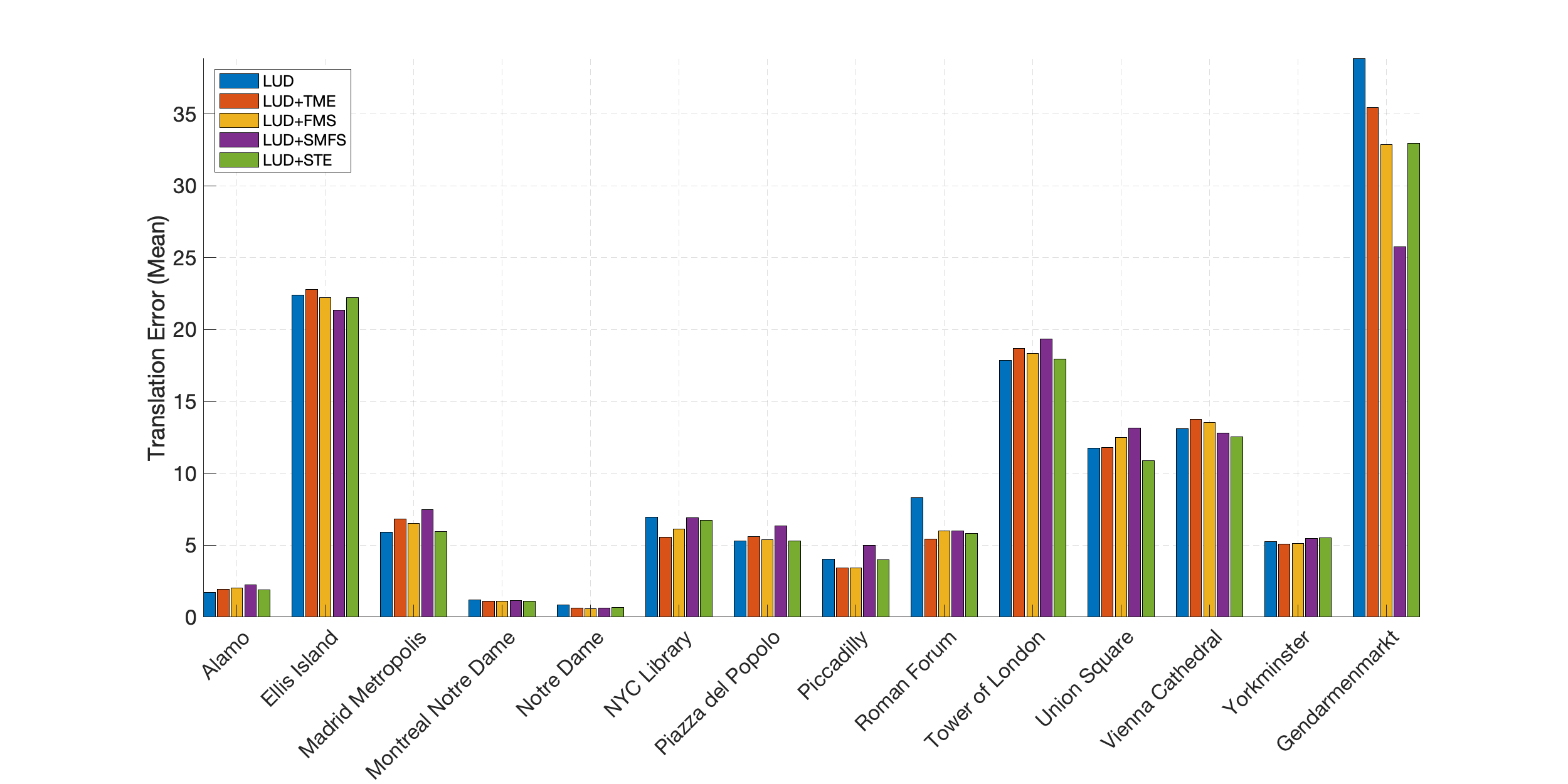}
    \caption{Median (top) and mean (bottom) absolute translation errors (in meters) of LUD and four RSR methods used to initially screen bad cameras within LUD applied to the 14 datasets of Photo Tourism.}
    \label{fig:remove_camera_2}
\end{figure*}

\begin{figure*}[htbp]
    \centering
    \includegraphics[width=0.9\textwidth]{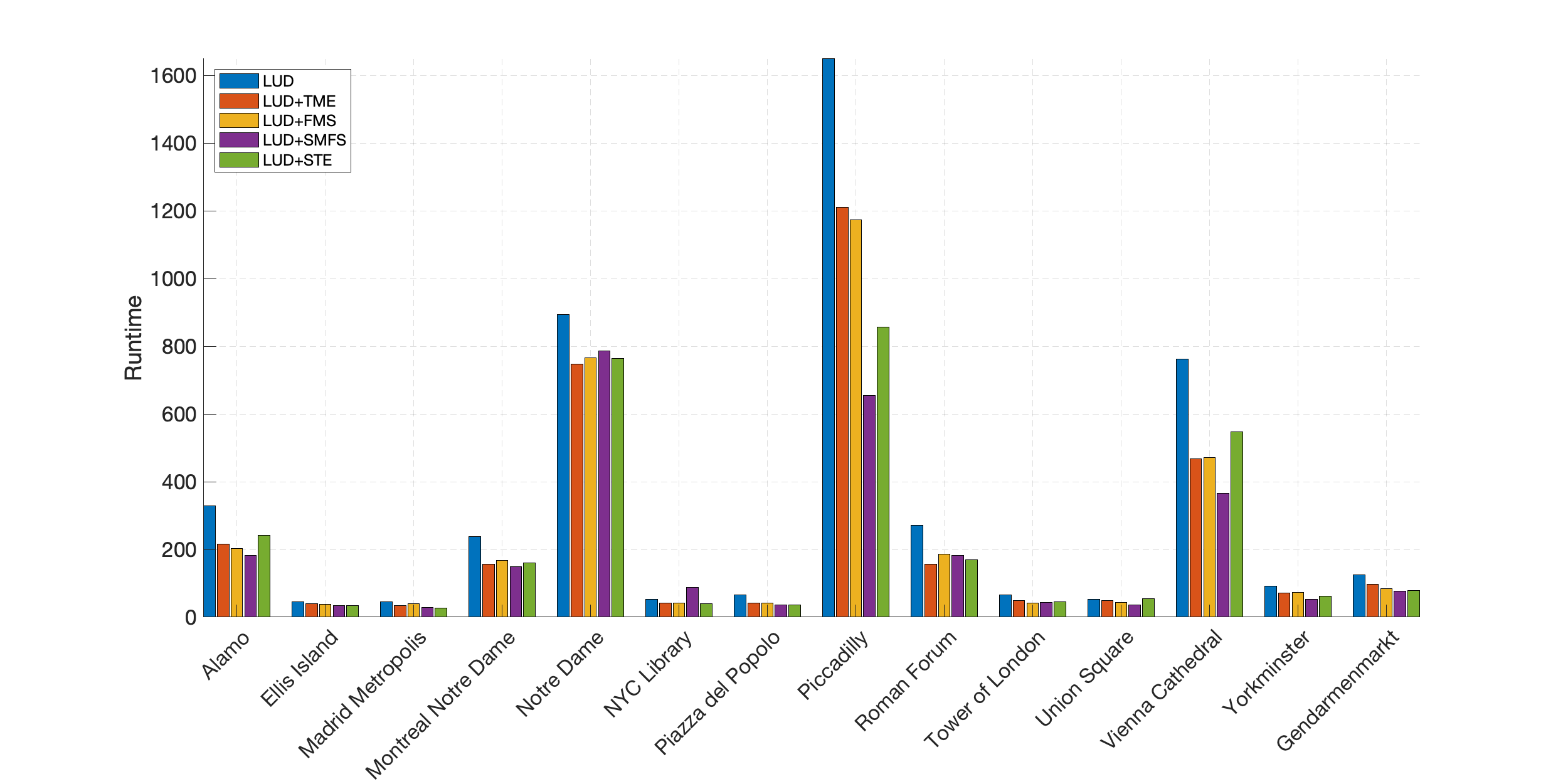}
    \caption{Runtime (in seconds) of the LUD pipeline and four LUD-type pipelines including an RSR method to initially screen bad cameras, applied to the 14 datasets of Photo Tourism.}
    \label{fig:remov_camera_runtime}
\end{figure*}

We compare mean and median errors of rotations and translations and runtime for 
the LUD pipeline and the LUD+RSR pipeline for the following four RSR algorithms: STE, FMS, SFMS, and TME. By LUD+RSR we refer to our proposed camera filtering process by the chosen RSR algorithm followed by LUD. 
\Cref{fig:remove_camera_1} 
is a bar plot of rotation errors (in degrees) for all methods and all 14 datasets, \Cref{fig:remove_camera_2}
is a bar plot of translation errors (in meters), and \Cref{fig:remov_camera_runtime} is a bar plot of run times. As explained in the main text, we only aim to compare the effect of the camera screening on the overall time (and this screening may potentially increase time due to issues with parallel rigidity), but it is not the actual time as LUD+RSR uses information for scaling factors obtained by LUD.  
For completeness, 
\Cref{tab:remove_camera_1} summarizes the results of LUD, LUD+STE, and LUD+SFMS, whereas \Cref{tab:remove_camera_2} presents the results of LUD+TME and LUD+FMS.

Overall, STE yields better accuracy than other RSR methods. 
We observe that LUD+STE generally improves the estimation of camera parameters (both rotations and translations) over LUD, though the median rotation errors are comparable. The improvement of LUD+STE is significant for both the Roman Forum and Gendarmenmarkt. In terms of runtime, both LUD+STE and LUD+SFMS demonstrate improvements, where LUD+SFMS is even faster than LUD+STE. While this does not yet imply faster handling of the datasets (as we use initial scaling factors obtained by LUD), it indicates some efficiency in the removal of outliers. 

\Cref{tab:gen-45} reports results for Gendarmenmarkt when removing $45\%$ of outlying columns. We note that STE significantly reduces the pipeline's runtime and improves the accuracy for both rotations and translations. TME also exhibits improvement, but it is less accurate and is also slower than STE.
On the other hand, FMS and SFMS do not improve the accuracy.
While the resulting errors are still large, their improvement shows some potential in dealing with difficult SfM structure by initially removing cameras in a way that may help eliminate some scene ambiguities, which are prevalent in Gendarmenmarkt.

\newpage

\begin{table*}[htbp]
\resizebox{\linewidth}{!}{%
\begin{tabular}{l|rr|rrrrrr|rrrrrr|rrrrrr}
\toprule
& & & & \multicolumn{4}{c}{LUD} & & \multicolumn{5}{c}{LUD+STE} & & \multicolumn{5}{c}{LUD+SFMS} \\
\cmidrule(lr){4-9} \cmidrule(lr){10-15} \cmidrule(lr){16-21}
Location & $n$ & $N$ & $n$ & $\hat{e}_T$ & $\tilde{e}_T$ & $\hat{e}_R$ & $\tilde{e}_R$ & $T_{\text{total}}$ & $n$ & $\hat{e}_T$ & $\tilde{e}_T$ & $\hat{e}_R$ & $\tilde{e}_R$ & $T_{\text{total}}$ & $n$ & $\hat{e}_T$ & $\tilde{e}_T$ & $\hat{e}_R$ & $\tilde{e}_R$ & $T_{\text{total}}$ \\
\midrule
Alamo & 570 & 606,963 & 563 & 1.74 & 0.47 & 3.04 & 1.03 & 328.8 & 497 & 1.90 & 0.45 & 3.08 & 1.04 & 242.6 & 466 & 2.24 & 0.46 & 4.44 & 1.81 & 183.3 \\
Ellis Island & 230 & 178,324 & 227 & 22.41 & 22.10 & 2.14 & 0.71 & 44.6 & 203 & 22.22 & 20.62 & 1.73 & 0.61 & 34.9 & 194 & 21.35 & 17.30 & 2.28 & 0.77 & 34.7 \\
Madrid Metropolis & 330 & 187,790 & 319 & 5.93 & 1.84 & 3.56 & 0.93 & 45.0 & 281 & 5.97 & 2.05 & 3.21 & 0.95 & 27.3 & 288 & 7.47 & 2.33 & 4.91 & 1.59 & 28.5 \\
Montreal N.D. & 445 & 633,938 & 437 & 1.22 & 0.56 & 0.96 & 0.49 & 237.8 & 388 & 1.12 & 0.52 & 1.02 & 0.52 & 159.7 & 381 & 1.15 & 0.54 & 1.05 & 0.49 & 149.5 \\
Notre Dame & 547 & 1,345,766 & 543 & 0.85 & 0.29 & 2.15 & 0.66 & 894.0 & 473 & 0.69 & 0.25 & 1.68 & 0.63 & 763.7 & 480 & 0.62 & 0.25 & 1.60 & 0.65 & 785.6 \\
NYC Library & 313 & 259,302 & 309 & 6.95 & 2.42 & 2.41 & 1.19 & 53.7 & 276 & 6.76 & 2.31 & 2.58 & 1.27 & 39.5 & 237 & 6.93 & 4.27 & 3.09 & 1.55 & 87.7 \\
Piazza del Popolo & 307 & 15,791 & 298 & 5.29 & 1.67 & 2.90 & 1.64 & 65.2 & 258 & 5.30 & 1.60 & 3.06 & 1.81 & 36.6 & 245 & 6.34 & 1.70 & 3.33 & 2.04 & 36.2 \\
Piccadilly & 2,226 & 1,278,612 & 2,171 & 4.04 & 1.81 & 4.82 & 1.72 & 1,649.0 & 1,937 & 3.98 & 1.95 & 4.37 & 1.76 & 857.4 & 1,827 & 4.98 & 2.72 & 5.33 & 2.01 & 654.4 \\
Roman Forum & 995 & 890,945 & 967 & 8.32 & 2.20 & 2.04 & 1.38 & 272.2 & 810 & 5.84 & 2.19 & 2.47 & 1.49 & 170.4 & 811 & 5.98 & 2.23 & 2.28 & 1.48 & 182.9 \\
Tower of London & 440 & 474171 & 431 & 17.86 & 3.96 & 2.92 & 2.13 & 65.2 & 355 & 17.95 & 4.79 & 3.01 & 2.63 & 44.9 & 338 & 19.35 & 4.62 & 3.64 & 2.92 & 43.1 \\
Union Square & 733 & 323933 & 715 & 11.75 & 7.57 & 6.59 & 3.21 & 53.7 & 651 & 10.90 & 7.29 & 6.21 & 3.47 & 53.8 & 571 & 13.14 & 8.81 & 6.93 & 3.92 & 36.5 \\
Vienna Cathedral & 789 & 1,361,659 & 774 & 13.10 & 7.21 & 7.28 & 1.65 & 762.6 & 685 & 12.56 & 6.28 & 7.91 & 1.81 & 547.1 & 666 & 12.82 & 7.78 & 8.08 & 1.80 & 365.6 \\
Yorkminster & 412 & 525,592 & 405 & 5.25 & 2.51 & 2.23 & 1.43 & 91.9 & 359 & 5.51 & 2.32 & 2.21 & 1.48 & 62.9 & 335 & 5.49 & 2.66 & 2.40 & 1.60 & 52.2 \\
Gendarmenmarkt & 671 & 338,800 & 654 & 38.82 & 17.46 & 39.63 & 6.43 & 124.2 & 596 & 32.94 & 14.70 & 29.01 & 4.65 & 79.2 & 564 & 25.77 & 15.88 & 38.16 & 5.50 & 77.8
\\
\bottomrule
\end{tabular}%
}
\caption{Performance of the LUD, LUD+STE and LUD+SFMS pipelines on the Photo Tourism datasets: $n$ and $N$ are the number of cameras and key points, respectively (when applying screening, $n$ also denotes the number of the remaining cameras); $\hat{e}_R,\tilde{e}_R$ indicate mean and median errors of absolute camera rotations in degrees, respectively; $\hat{e}_T,\tilde{e}_T$ indicate mean and median errors of absolute camera translations in meters, respectively; $T_{\text{total}}$ is the runtime of the pipeline (in seconds). }
\label{tab:remove_camera_1}
\end{table*}

\begin{table*}[htbp]
\resizebox{\linewidth}{!}{%
\begin{tabular}{l|rr|rrrrrr|rrrrrr}
\toprule
& & & & \multicolumn{4}{c}{LUD+TME} & & \multicolumn{5}{c}{LUD+FMS} &\\
\cmidrule(lr){4-9} \cmidrule(lr){10-15} 
Location & $n$ & $N$ & $n$ & $\hat{e}_T$ & $\tilde{e}_T$ & $\hat{e}_R$ & $\tilde{e}_R$ & $T_{\text{total}}$ & $n$ & $\hat{e}_T$ & $\tilde{e}_T$ & $\hat{e}_R$ & $\tilde{e}_R$ & $T_{\text{total}}$ \\
\midrule
Alamo                & 570   & 606,963 & 463   & 1.96 & 0.48 & 3.04 & 1.07 & 216.6 & 480   & 2.05 & 0.51 & 3.26 & 1.10 & 203.0 \\
Ellis Island         & 230   & 178,324 & 205   & 22.80 & 22.01 & 2.87 & 0.93 & 39.6 & 207   & 22.22 & 20.60 & 2.75 & 0.90 & 38.4 \\
Madrid Metropolis    & 330   & 187,790 & 288   & 6.83 & 2.17 & 5.73 & 2.00 & 34.0 & 289   & 6.52 & 1.97 & 4.80 & 1.77 & 40.0 \\
Montreal N.D.  & 445   & 633,938 & 381   & 1.10 & 0.53 & 0.96 & 0.48 & 155.8 & 389   & 1.12 & 0.54 & 0.97 & 0.50 & 167.0 \\
Notre Dame           & 547   & 1,345,766 & 463   & 0.62 & 0.25 & 1.59 & 0.63 & 747.4 & 479   & 0.61 & 0.26 & 1.37 & 0.66 & 765.9 \\
NYC Library          & 313   & 259,302 & 267   & 5.55 & 1.91 & 2.53 & 1.15 & 41.6 & 263   & 6.14 & 2.03 & 2.52 & 1.25 & 42.1 \\
Piazza del Popolo    & 307   & 15,791 & 251   & 5.61 & 1.56 & 2.96 & 1.68 & 41.8 & 257   & 5.40 & 1.62 & 3.05 & 1.69 & 41.3 \\
Piccadilly           & 2,226 & 1,278,612 & 1,980 & 3.44 & 1.65 & 3.19 & 1.77 & 1,211.7 & 1,888 & 3.43 & 1.68 & 3.20 & 1.75 & 1,172.9 \\
Roman Forum          & 995   & 890,945 & 828   & 5.42 & 2.08 & 2.34 & 1.46 & 157.0 & 814   & 6.01 & 2.19 & 2.35 & 1.38 & 185.8 \\
Tower of London      & 440   & 474,171 & 364   & 18.67 & 4.93 & 3.02 & 2.39 & 48.8 & 359   & 18.34 & 4.62 & 3.02 & 2.62 & 40.9 \\
Union Square         & 733   & 323,933 & 612   & 11.80 & 7.68 & 6.18 & 3.41 & 49.4 & 610   & 12.49 & 7.48 & 6.13 & 3.37 & 44.1 \\
Vienna Cathedral     & 789   & 1,361,659 & 694   & 13.77 & 7.26 & 7.69 & 2.89 & 468.7 & 692   & 13.54 & 7.17 & 7.70 & 2.90 & 472.0 \\
Yorkminster          & 412   & 525,592 & 360   & 5.07 & 2.40 & 2.15 & 1.47 & 70.8 & 358   & 5.11 & 2.44 & 2.19 & 1.51 & 73.3 \\
Gendarmenmarkt       & 671   & 338,800 & 602   & 35.43 & 14.69 & 28.30 & 4.08 & 96.7 & 600   & 32.87 & 14.34 & 28.59 & 4.09 & 85.0 \\
\bottomrule
\end{tabular}%
}
\caption{Performance of the LUD+TME and LUD+FMS pipelines on the Photo Tourism datasets: $n$ and $N$ are the number of cameras and key points, respectively (when applying screening, $n$ also denotes the number of the remaining cameras); $\hat{e}_R,\tilde{e}_R$ indicate mean and median errors of absolute camera rotations in degrees, respectively; $\hat{e}_T,\tilde{e}_T$ indicate mean and median errors of absolute camera translations in meters, respectively; $T_{\text{total}}$ is the runtime of the pipeline (in seconds). }
\label{tab:remove_camera_2}
\end{table*}

\begin{table}[htbp]
\centering
\begin{tabular}{l|rrrrrr}
\toprule
Method   & $n$ & $\hat{e}_T$ & $\tilde{e}_T$ & $\hat{e}_R$ & $\tilde{e}_R$ & $T_{\text{total}}$ \\
\midrule
LUD      & 654 & 38.82 & 17.46 & 39.63 & 6.43 & 124.2 \\
LUD+STE  & 497 & 22.82 & 12.53 & 26.76 & 3.60 & 51.24 \\
LUD+TME  & 516 & 25.07 & 14.72 & 24.93 & 4.18 & 64.35 \\
LUD+FMS  & 559 & 40.74 & 18.6 & 40.28 & 6.54 & 102.14 \\
LUD+SFMS & 549 & 41.31 & 19.33 & 40.65 & 6.47 & 100.40 \\
\bottomrule
\end{tabular}
\caption{Comparison of the four pipelines when $45\%$ outlying columns are eliminated in Gendarmenmarkt. Recall that  $\hat{e}_R,\tilde{e}_R$ indicate mean and median errors of absolute camera rotations in degrees, respectively; $\hat{e}_T,\tilde{e}_T$ indicate mean and median errors of absolute camera translations in meters, respectively; and $T_{\text{total}}$ is the runtime of the pipeline (in seconds).}
\label{tab:gen-45}
\end{table}

\begin{figure*}[htbp]
    \includegraphics[width=0.24\textwidth]{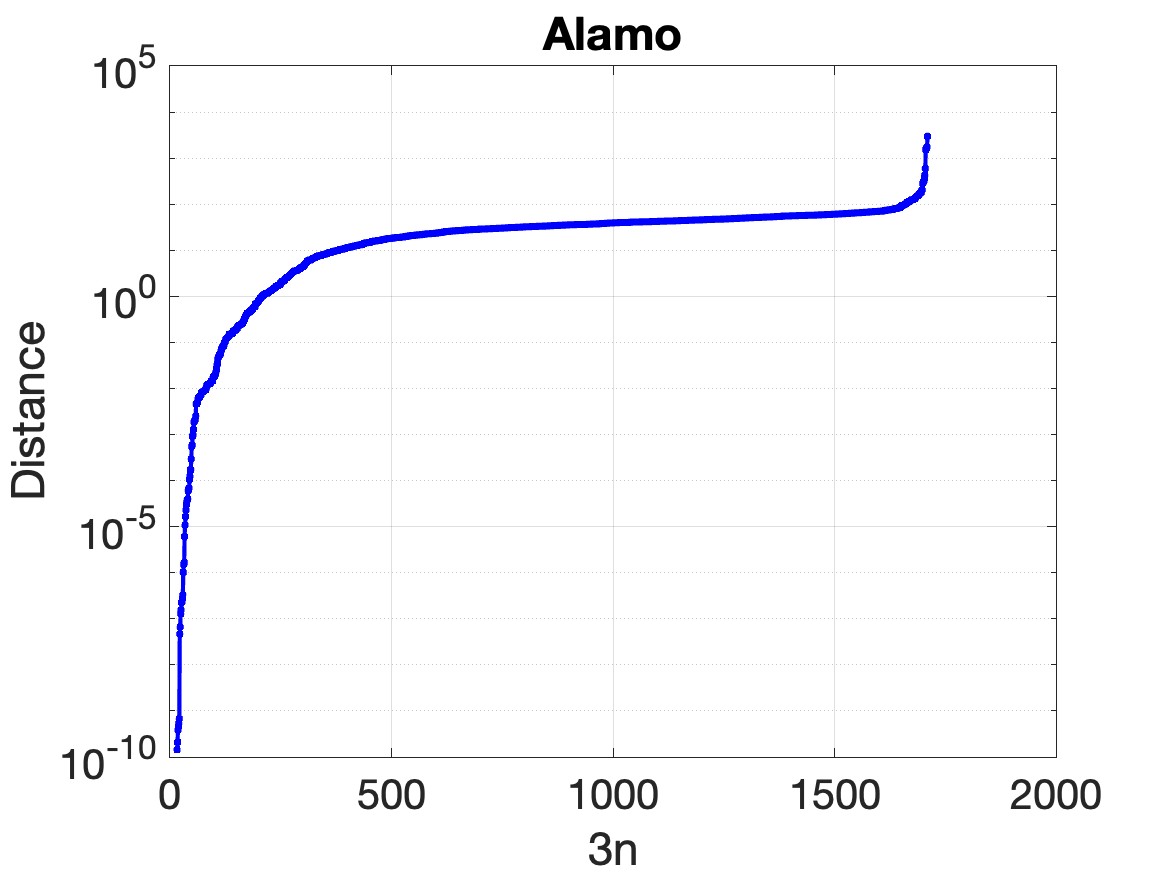}
    \includegraphics[width=0.24\textwidth]{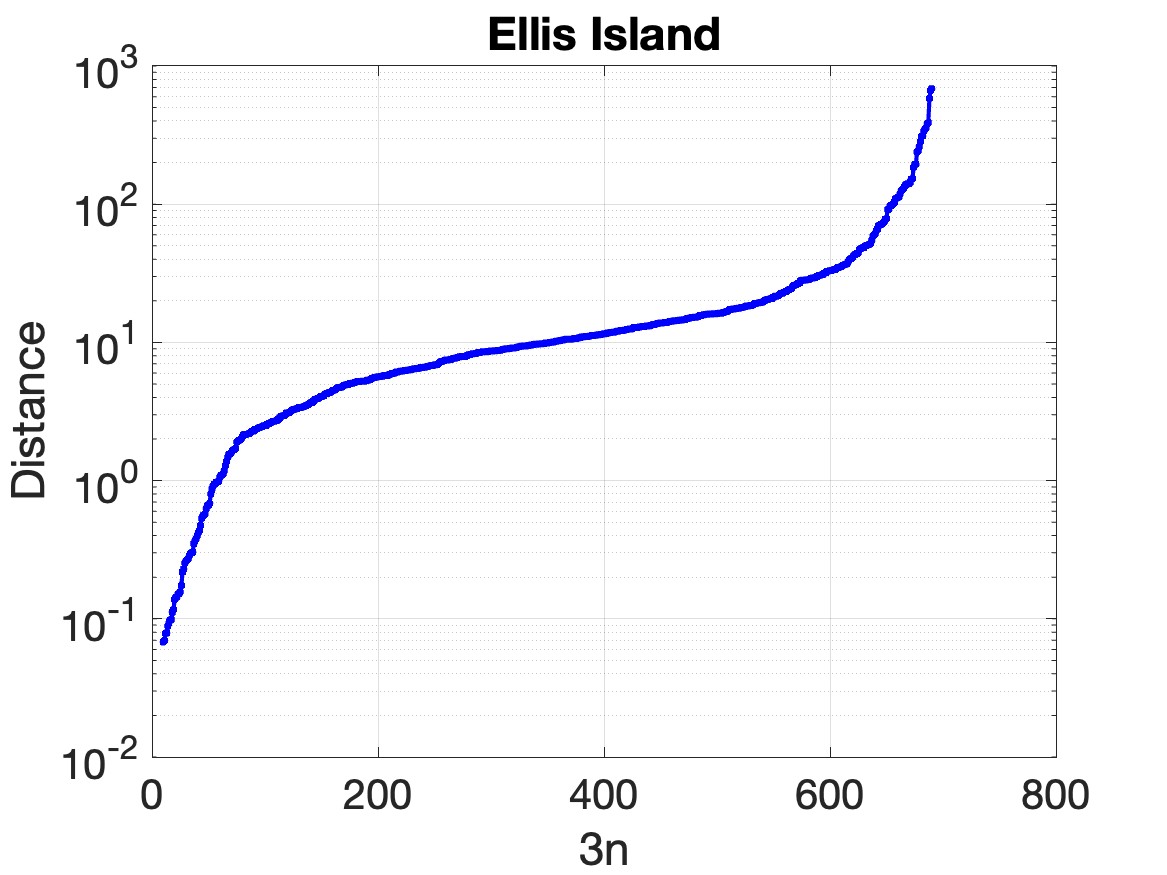}
    \includegraphics[width=0.24\textwidth]{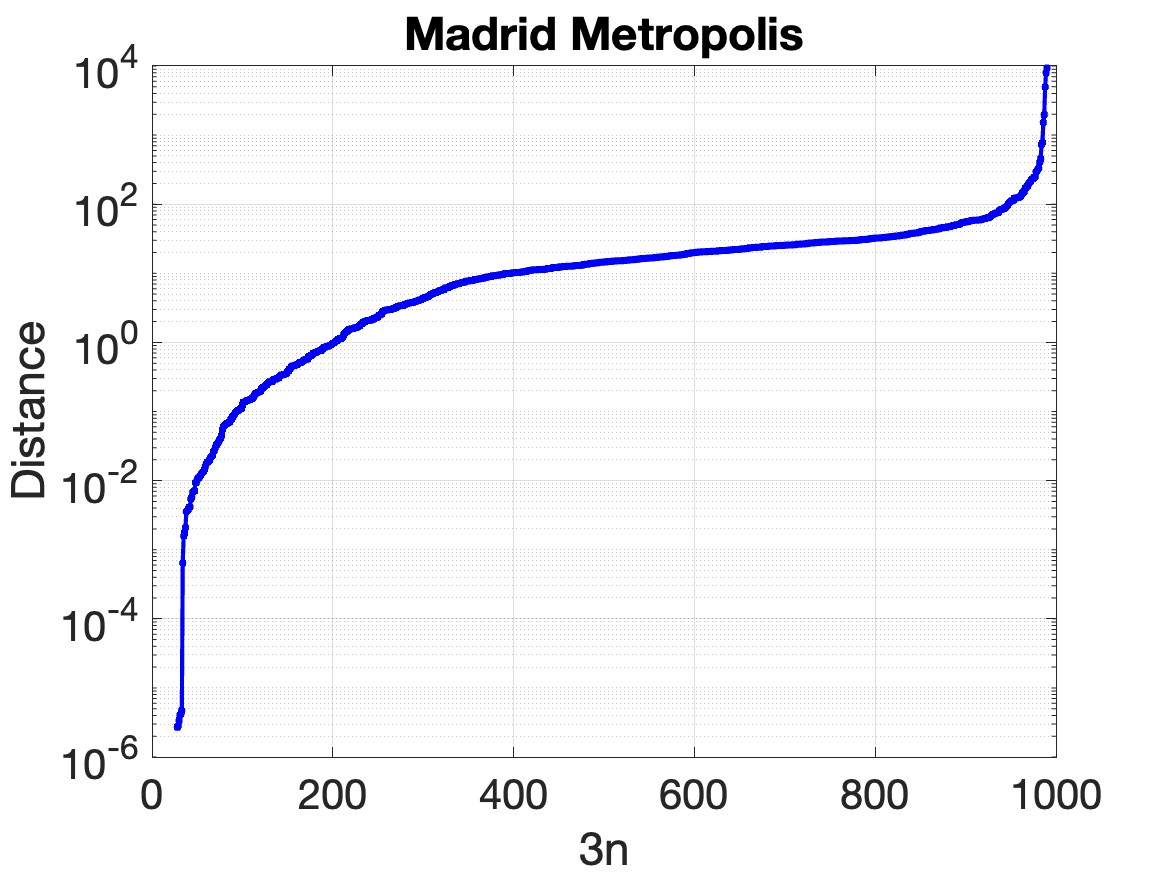}
    \includegraphics[width=0.24\textwidth]{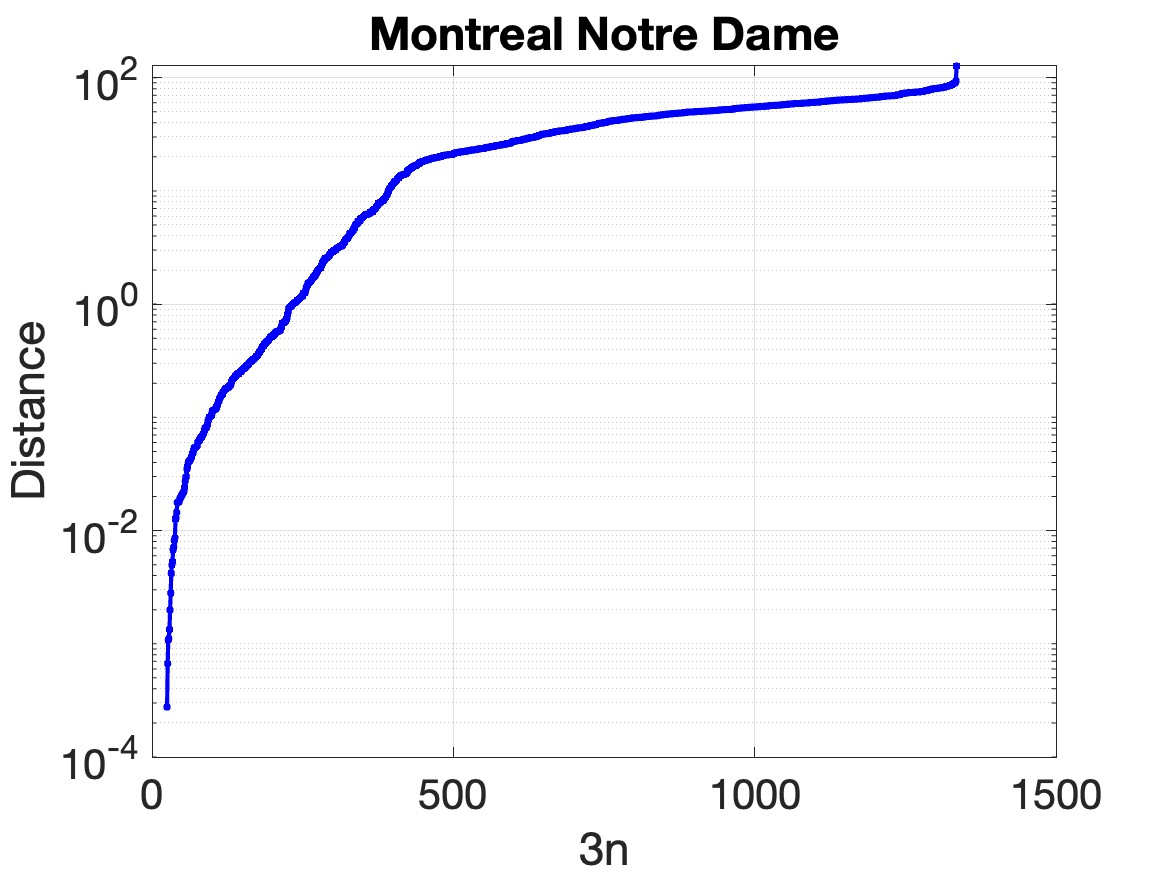}
    \includegraphics[width=0.24\textwidth]{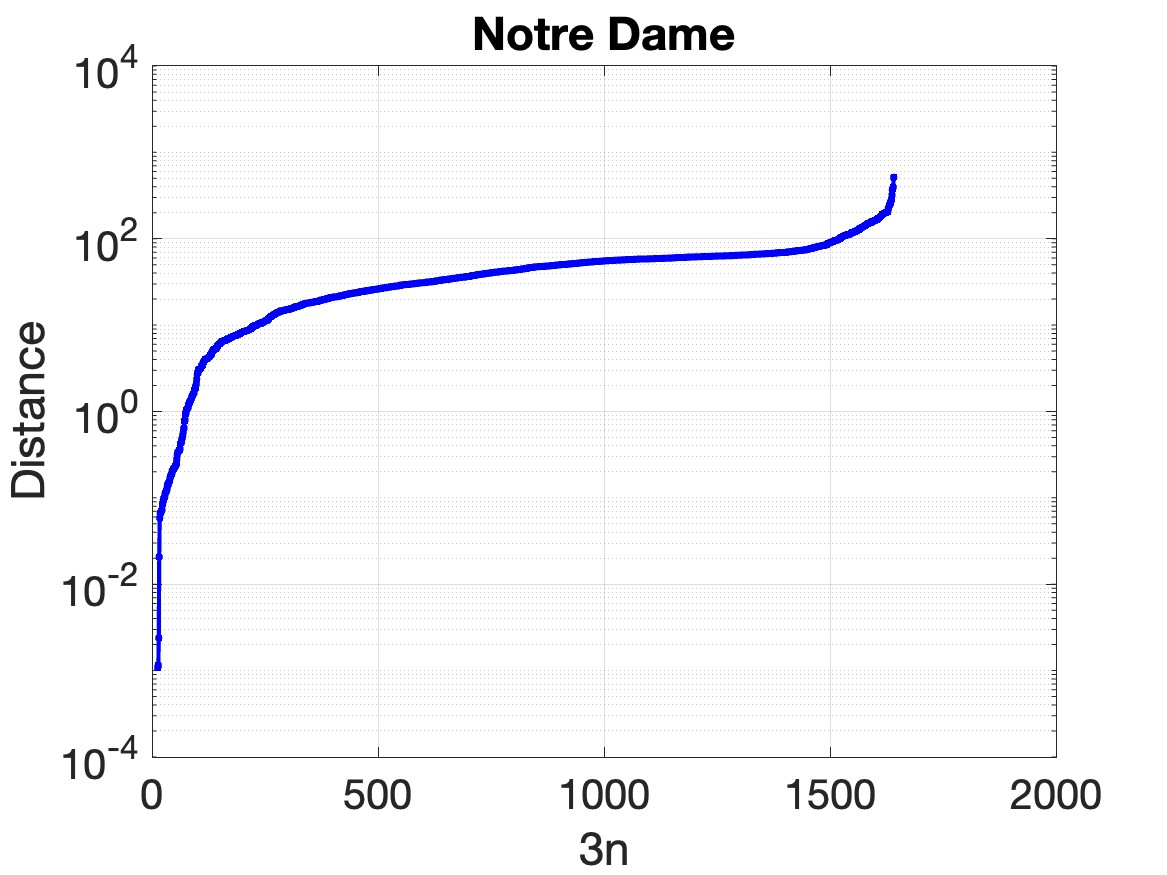}
    \includegraphics[width=0.24\textwidth]{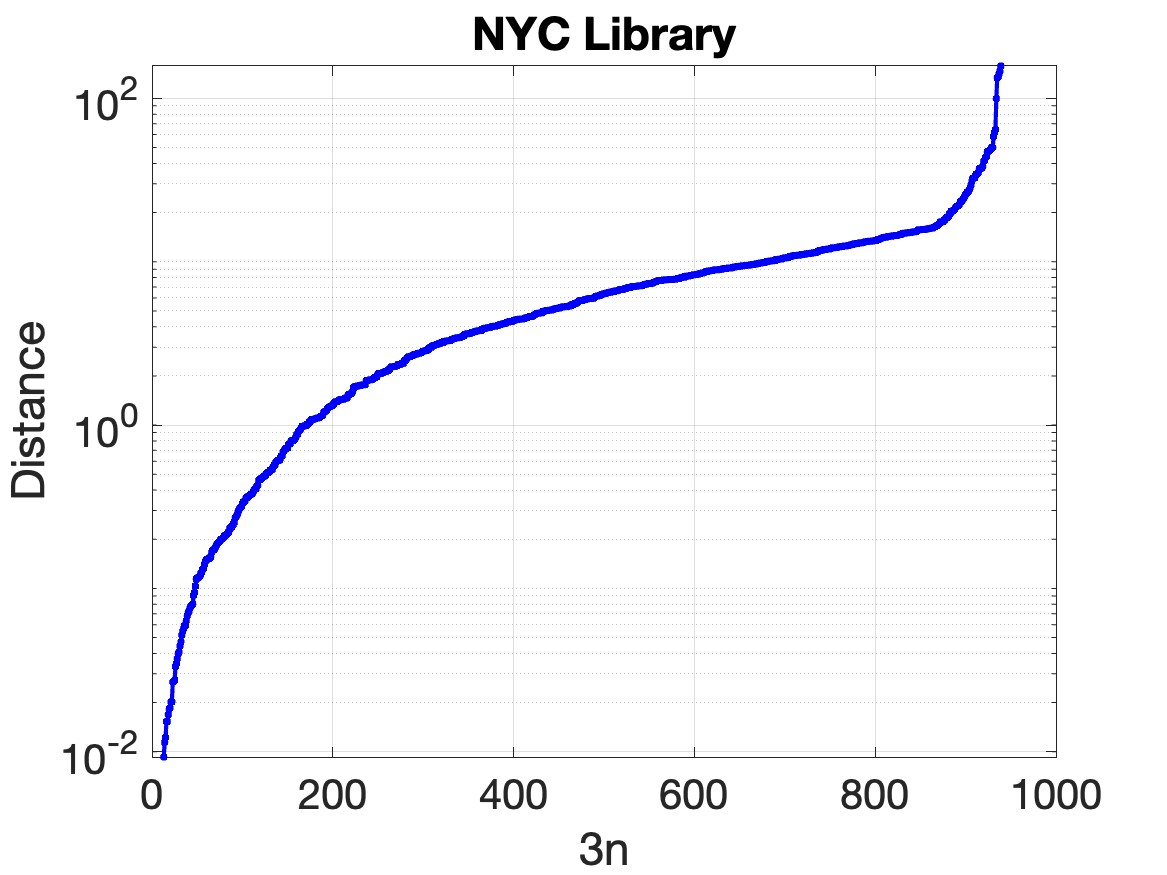}
    \includegraphics[width=0.24\textwidth]{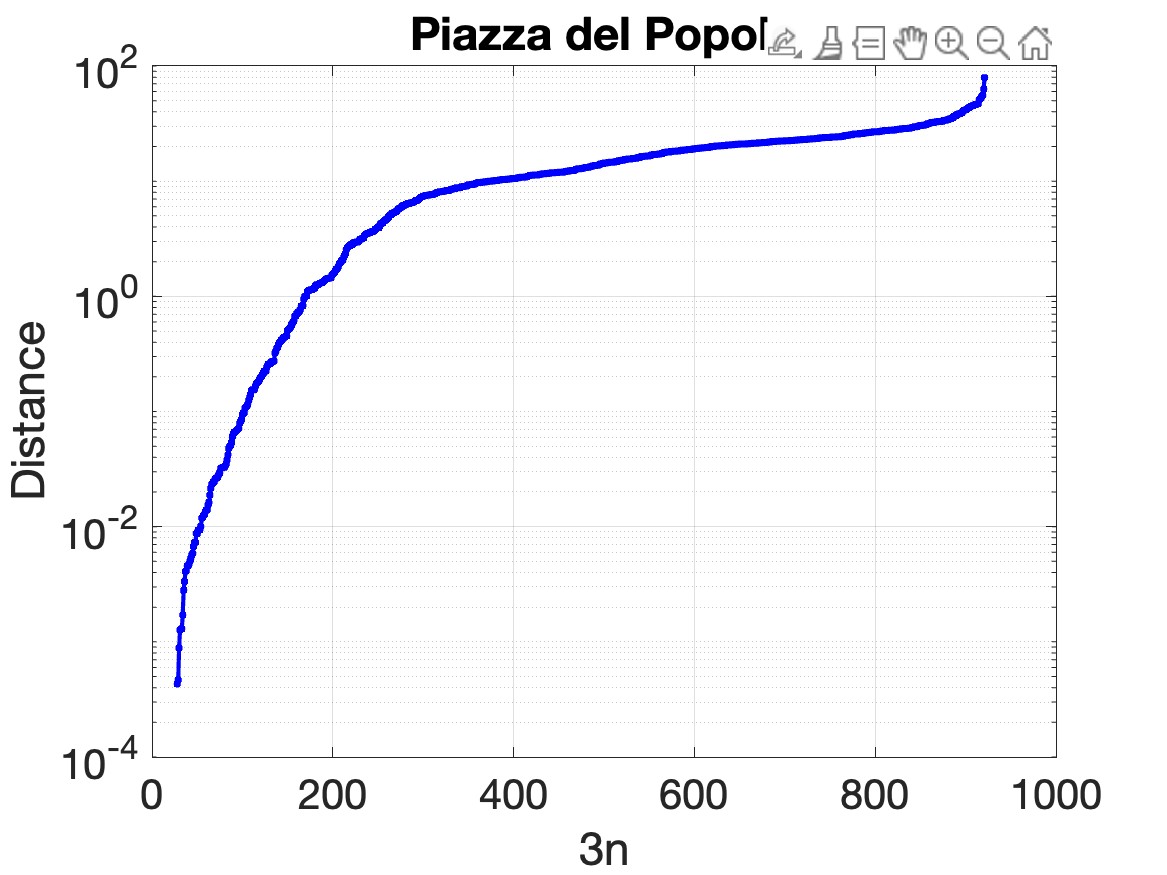}
    \includegraphics[width=0.24\textwidth]{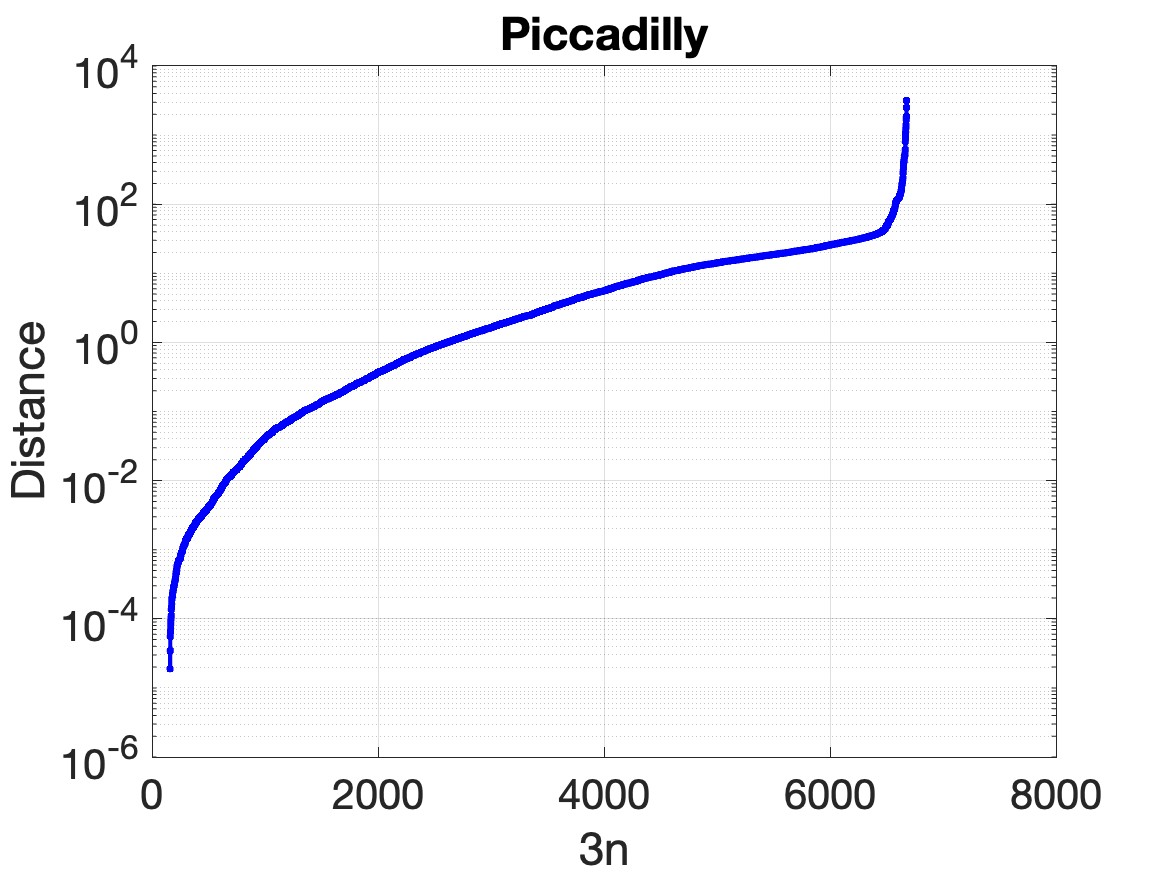}
    \includegraphics[width=0.24\textwidth]{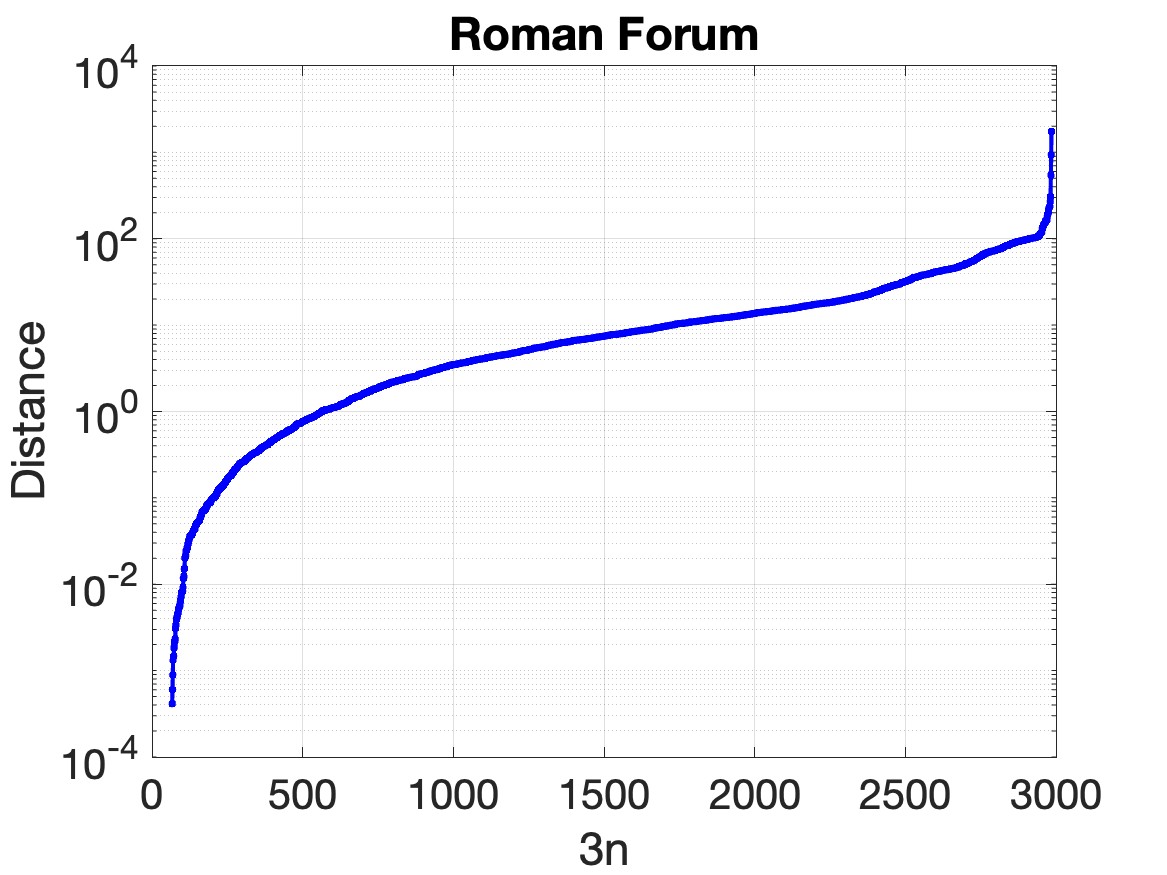}
    \includegraphics[width=0.24\textwidth]{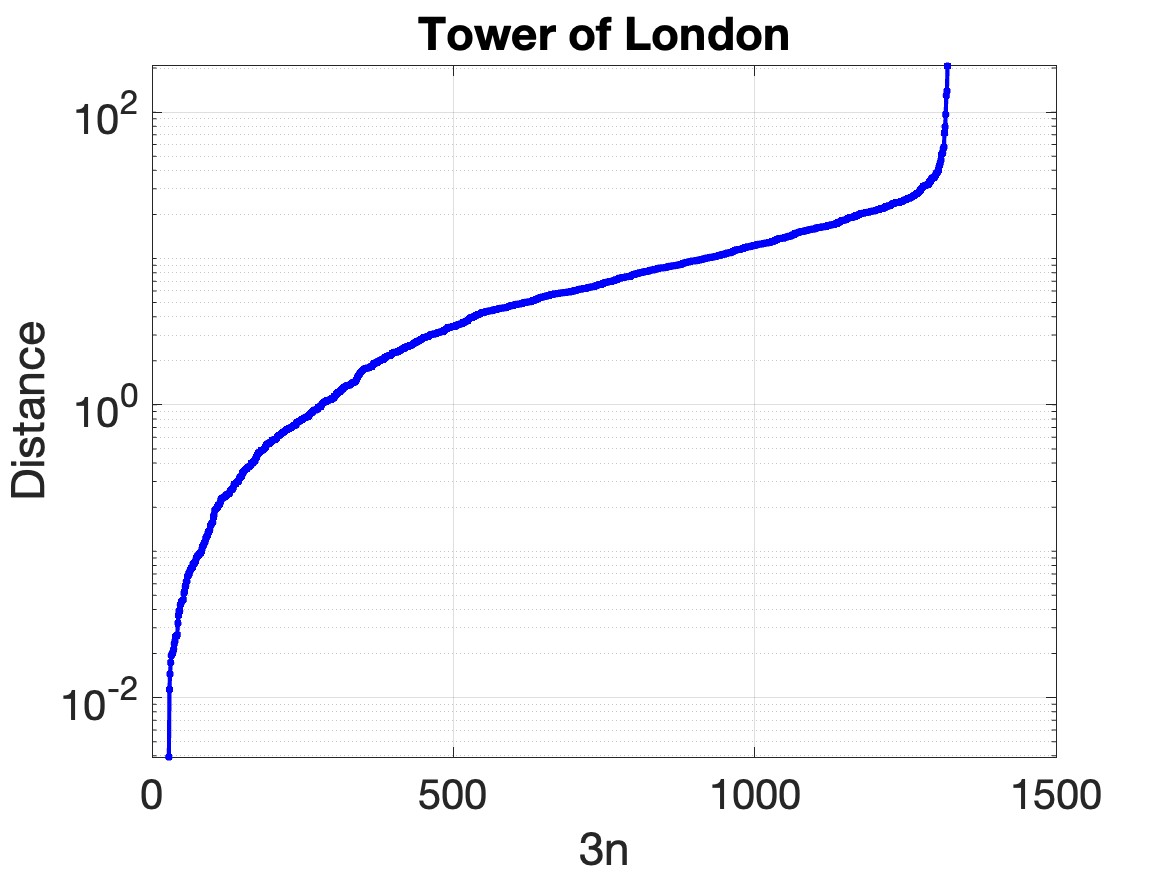}
    \includegraphics[width=0.24\textwidth]{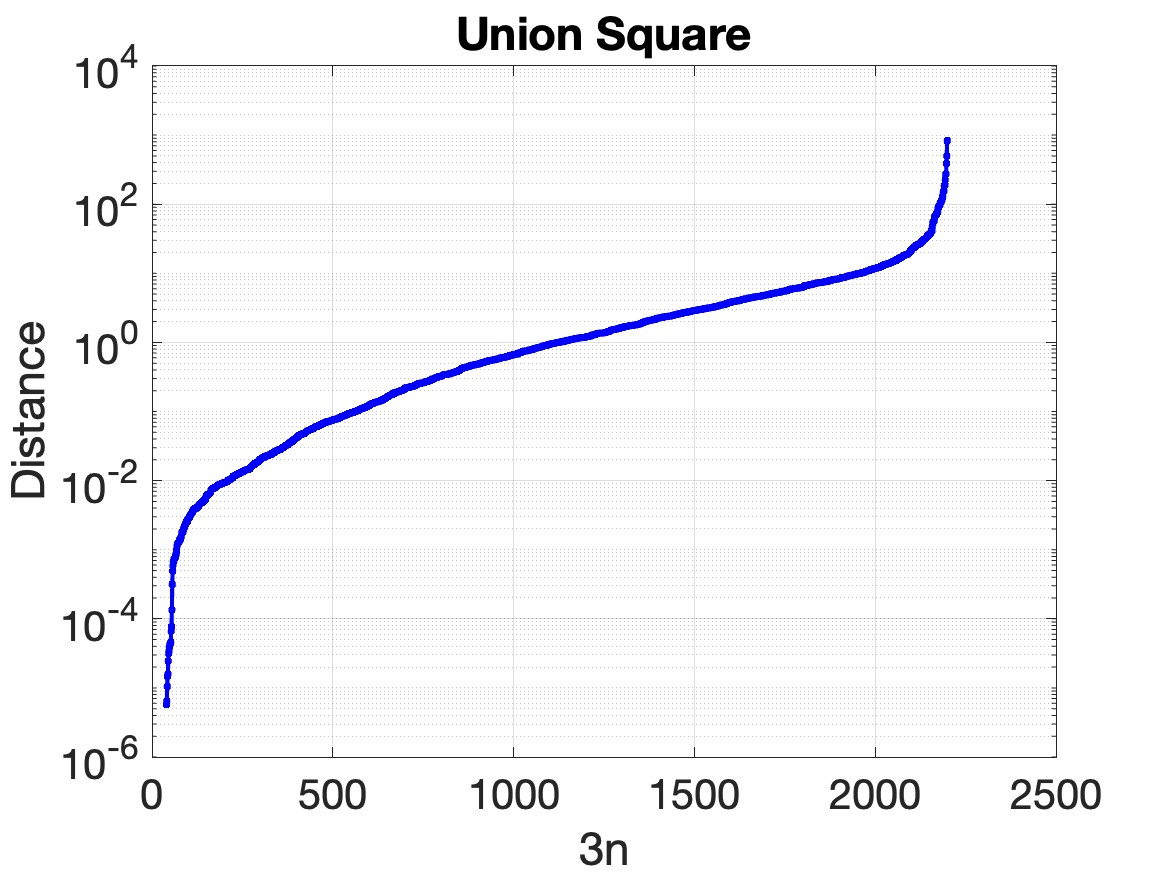}
    \includegraphics[width=0.24\textwidth]{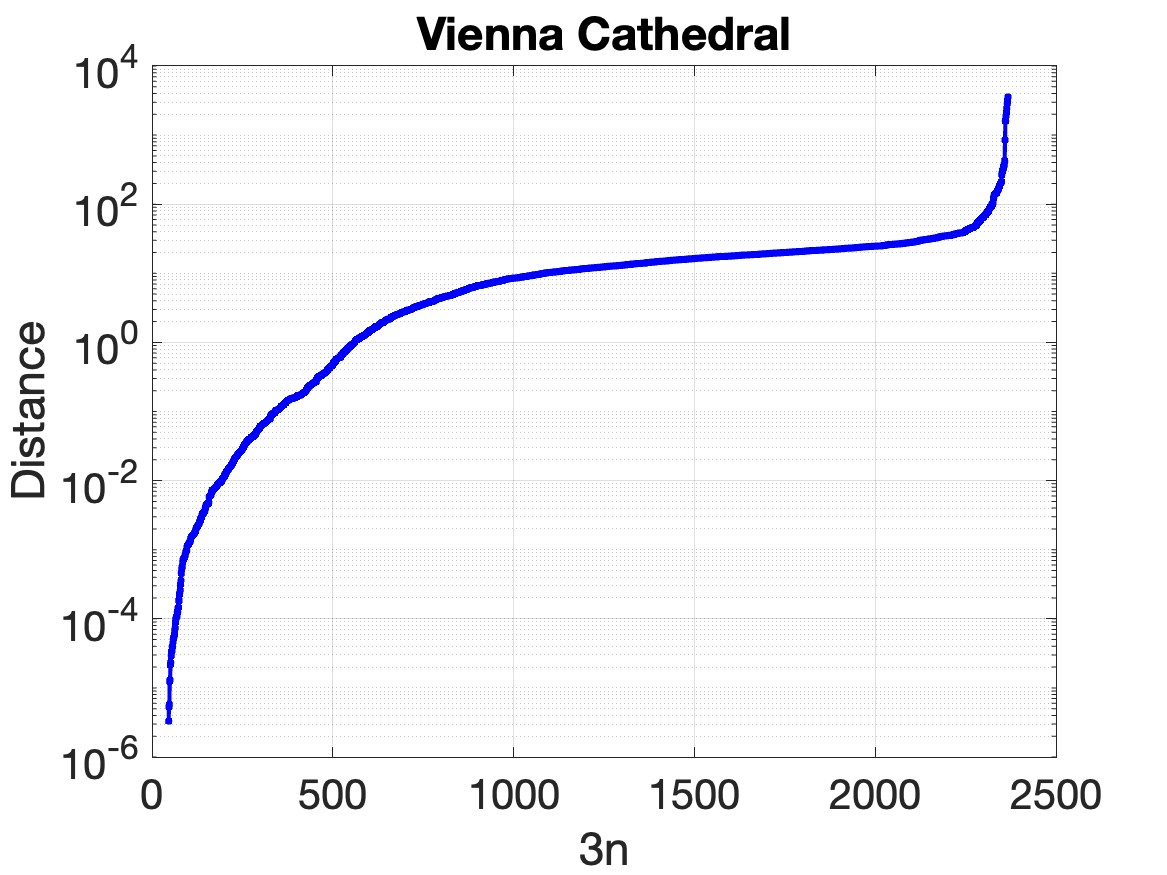}
    \includegraphics[width=0.24\textwidth]{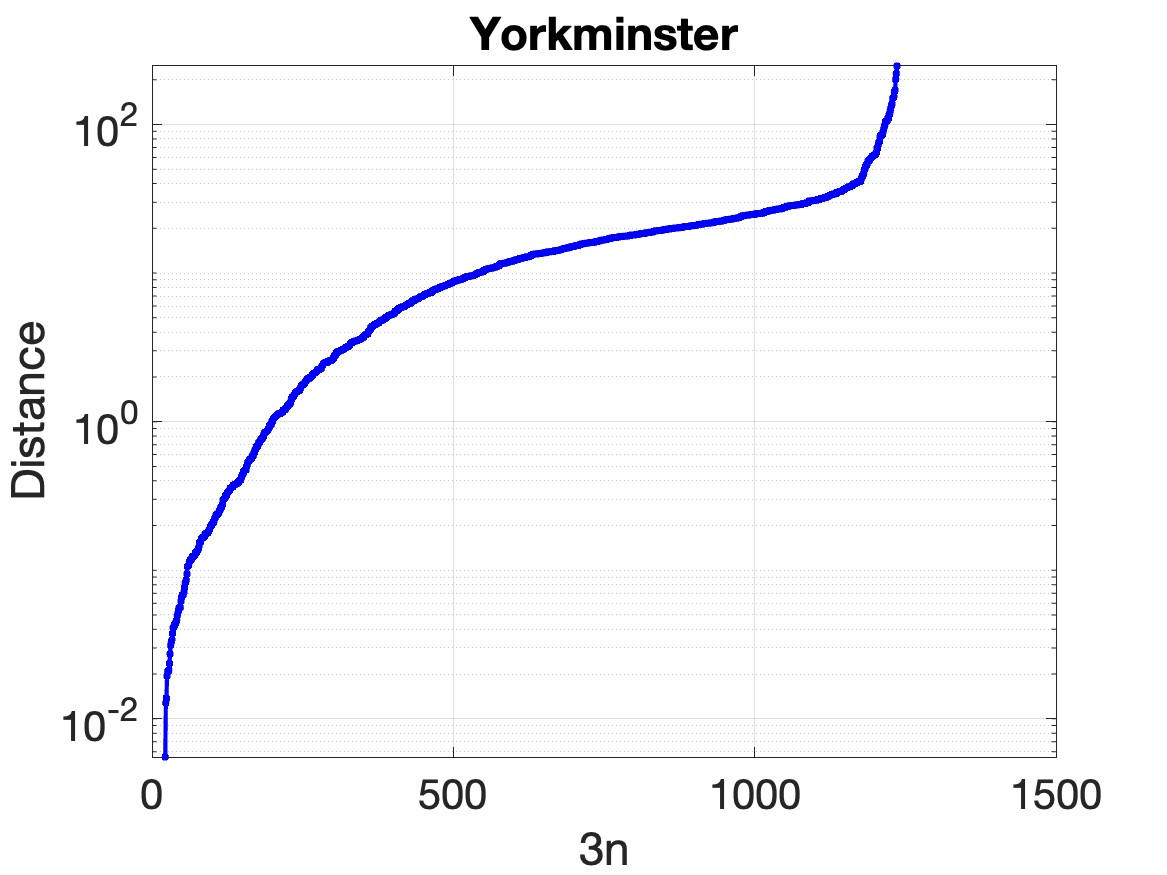}
    \includegraphics[width=0.24\textwidth]{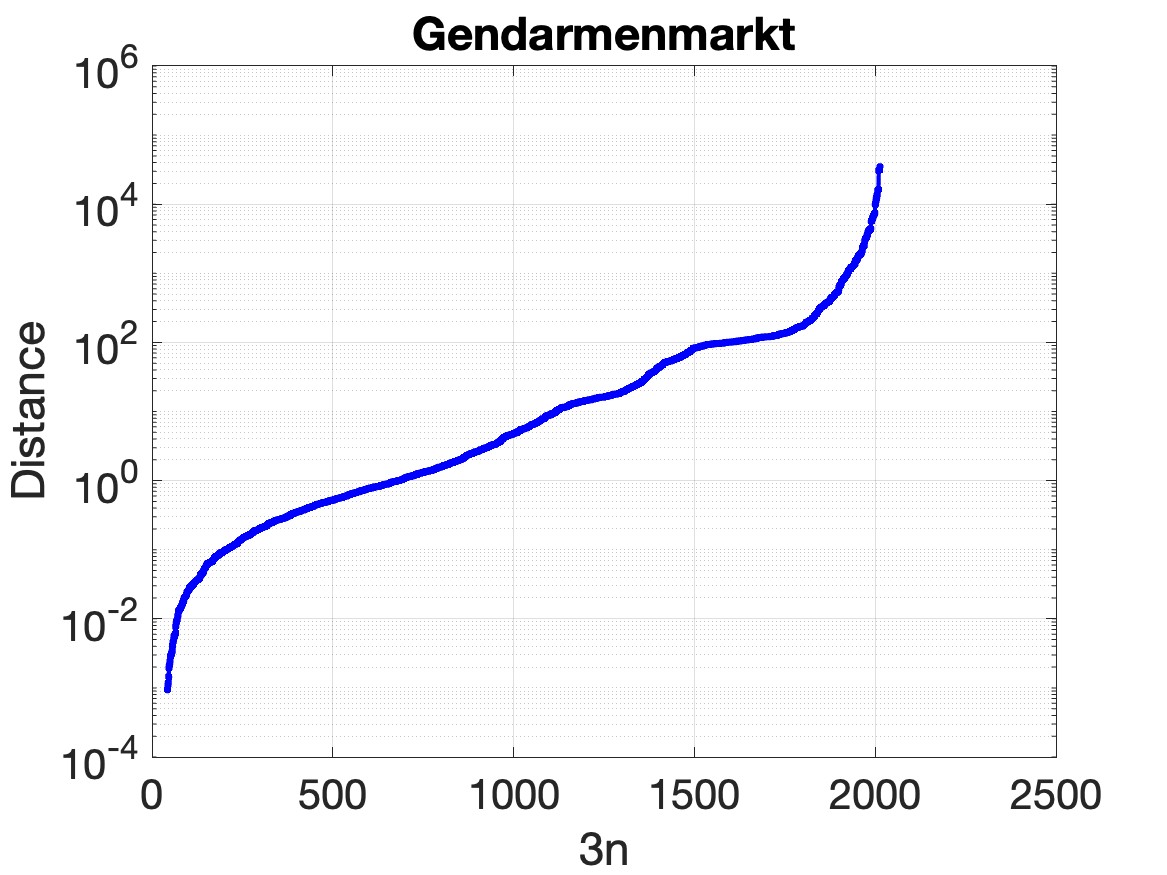}
    \caption{Distance of columns of ${\bE}$ to the subspace recovered by STE (in log scale). The distances are sorted and plotted in descending order. All datasets exhibit significant outlying columns.}
    \label{fig:dist_SfM}
\end{figure*}

Finally, in order to get an idea about the behavior of the outliers, \Cref{fig:dist_SfM}
plots the distances of all data points to the STE subspace, where it uses a logarithmic scale.  Distinguished outliers are noticed in all datasets, but Piazza del Popolo. Moreover, such significant outliers constitute only a small portion of the columns.  As mentioned in the main text, we preferred to avoid heuristic methods for the cutoff of outliers, and thus assumed a fixed percentage of $20\%$ outlying column vectors.

\newpage
\subsubsection{Initial Camera Removal for SfM without Matrix Completion}
\label{sec:sfm_no_mc}
For completeness, we also provide the numerical results for the initial camera removal experiment when the absent blocks $\bE_{ij}$ are assigned zero matrices. \Cref{tab:remove_camera_1_w/t_MC} and \Cref{tab:remove_camera_2_w/t_MC} showcase the performance of the LUD+RSR piplines without the matrix completion step. Overall, integrating matrix completion into the LUD+RSR pipelines enhances their performance. The most notable enhancement for STE is for the mean translation error of the Roman Forum and Gendarmenmarkt. For other RSR algorithm we notice even more significant improvement with matrix completion. In other words, without matrix completion, the higher accuracy of STE in comparison to other RSR methods is more noticeable.

\begin{table*}[htbp]
\resizebox{\linewidth}{!}{%
\begin{tabular}{l|rr|rrrrrr|rrrrrr|rrrrrr}
\toprule
& & & & \multicolumn{4}{c}{LUD} & & \multicolumn{5}{c}{LUD+STE} & & \multicolumn{5}{c}{LUD+SFMS} \\
\cmidrule(lr){4-9} \cmidrule(lr){10-15} \cmidrule(lr){16-21}
Location & $n$ & $N$ & $n$ & $\hat{e}_T$ & $\tilde{e}_T$ & $\hat{e}_R$ & $\tilde{e}_R$ & $T_{\text{total}}$ & $n$ & $\hat{e}_T$ & $\tilde{e}_T$ & $\hat{e}_R$ & $\tilde{e}_R$ & $T_{\text{total}}$ & $n$ & $\hat{e}_T$ & $\tilde{e}_T$ & $\hat{e}_R$ & $\tilde{e}_R$ & $T_{\text{total}}$ \\
\midrule
Alamo                 & 570   & 606,963   & 563  & 1.74  & 0.47  & 3.04  & 1.03 & 328.8   & 518  & 1.90  & 0.45  & 3.09  & 1.06 & 248.6 & 466  & 2.24  & 0.47  & 4.44  & 1.82  & 219.0 \\
Ellis Island         & 230   & 178,324   & 227  & 22.41 & 22.10 & 2.14  & 0.71 & 44.6    & 209  & 22.17 & 20.45 & 1.74  & 0.61 & 39.3  & 195  & 21.51 & 18.50 & 2.27  & 0.77  & 33.3  \\
Madrid Metropolis    & 330   & 187,790   & 319  & 5.93  & 1.84  & 3.56  & 0.93 & 45.0    & 290  & 5.88  & 1.90  & 3.25  & 0.97 & 33.6  & 278  & 5.82  & 2.25  & 3.54  & 1.12  & 32.1  \\
Montreal Notre Dame & 445   & 633,938   & 437  & 1.22  & 0.56  & 0.96  & 0.49 & 237.8   & 400  & 1.12  & 0.55  & 1.18  & 0.56 & 149.9 & 383  & 1.14  & 0.54  & 1.03  & 0.53  & 146.5 \\
Notre Dame           & 547   & 1,345,766 & 543  & 0.85  & 0.29  & 2.15  & 0.66 & 894.0   & 495  & 0.68  & 0.26  & 1.65  & 0.74 & 763.8 & 476  & 0.72  & 0.26  & 1.87  & 0.66  & 815.6 \\
NYC Library          & 313   & 259,302   & 309  & 6.95  & 2.42  & 2.41  & 1.19 & 53.7    & 278  & 6.07  & 2.67  & 2.54  & 1.29 & 33.0  & 238  & 6.82  & 4.61  & 3.13  & 1.49  & 42.2  \\
Piazza del Popolo   & 307   & 15,791    & 298  & 5.29  & 1.67  & 2.90  & 1.64 & 65.2    & 247  & 2.40  & 1.33  & 1.26  & 0.58 & 27.9  & 242  & 6.62  & 1.59  & 3.51  & 2.15  & 45.9  \\
Piccadilly            & 2,226 & 1,278,612 & 2171 & 4.04  & 1.81  & 4.82  & 1.72 & 1,649.0 & 1976 & 4.35  & 2.08  & 4.78  & 1.62 & 842.7 & 1827 & 4.98  & 2.72  & 5.33  & 2.01  & 661.6 \\
Roman Forum          & 995   & 890,945   & 967  & 8.32  & 2.20  & 2.04  & 1.38 & 272.2   & 849  & 9.25  & 2.02  & 2.12  & 1.39 & 216.7 & 814  & 9.46  & 2.68  & 2.15  & 1.50  & 150.5 \\
Tower of London     & 440   & 474,171   & 431  & 17.86 & 3.96  & 2.92  & 2.13 & 65.2    & 373  & 19.84 & 2.85  & 2.66  & 1.95 & 29.8  & 351  & 22.31 & 4.10  & 3.18  & 2.46  & 27.0  \\
Union Square         & 733   & 323,933   & 715  & 11.75 & 7.57  & 6.59  & 3.21 & 53.7    & 624  & 12.64 & 8.47  & 7.11  & 3.95 & 39.9  & 558  & 13.42 & 9.04  & 6.87  & 3.48  & 38.5  \\
Vienna Cathedral     & 789   & 1,361,659 & 774  & 13.10 & 7.21  & 7.28  & 1.65 & 762.6   & 715  & 12.45 & 6.38  & 7.84  & 1.85 & 499.7 & 666  & 12.82 & 7.78  & 8.08  & 1.80  & 444.0 \\
Yorkminster           & 412   & 525,592   & 405  & 5.25  & 2.51  & 2.23  & 1.43 & 91.9    & 368  & 4.08  & 2.39  & 2.25  & 1.48 & 56.1  & 338  & 5.67  & 3.13  & 2.41  & 1.56  & 48.0  \\
Gendarmenmarkt        & 671   & 338,800   & 654  & 38.82 & 17.46 & 39.63 & 6.43 & 124.2   & 616  & 35.37 & 14.68 & 27.99 & 4.08 & 89.5  & 543  & 35.34 & 27.53 & 62.89 & 31.04 & 83.7 
\\
\bottomrule
\end{tabular}%
}
\caption{Performance of the LUD, LUD+STE and LUD+SFMS pipelines \textbf{without matrix completion} on the Photo Tourism datasets: $n$ and $N$ are the number of cameras and key points, respectively (when applying screening, $n$ also denotes the number of the remaining cameras); $\hat{e}_R,\tilde{e}_R$ indicate mean and median errors of absolute camera rotations in degrees, respectively; $\hat{e}_T,\tilde{e}_T$ indicate mean and median errors of absolute camera translations in meters, respectively; $T_{\text{total}}$ is the runtime of the pipeline (in seconds). }
\label{tab:remove_camera_1_w/t_MC}
\end{table*}

\begin{table*}[htbp]
\resizebox{\linewidth}{!}{%
\begin{tabular}{l|rr|rrrrrr|rrrrrr}
\toprule
& & & & \multicolumn{4}{c}{LUD+TME} & & \multicolumn{5}{c}{LUD+FMS} &\\
\cmidrule(lr){4-9} \cmidrule(lr){10-15} 
Location & $n$ & $N$ & $n$ & $\hat{e}_T$ & $\tilde{e}_T$ & $\hat{e}_R$ & $\tilde{e}_R$ & $T_{\text{total}}$ & $n$ & $\hat{e}_T$ & $\tilde{e}_T$ & $\hat{e}_R$ & $\tilde{e}_R$ & $T_{\text{total}}$ \\
\midrule
Alamo                 & 570   & 606,963   & 463  & 1.94  & 0.49  & 3.05  & 1.07 & 220.2   & 480  & 2.05  & 0.51  & 3.27  & 1.10 & 316.9   \\
Ellis Island         & 230   & 178,324   & 205  & 22.44 & 21.66 & 2.23  & 0.74 & 39.8    & 207  & 22.22 & 20.60 & 2.75  & 0.90 & 50.5    \\
Madrid Metropolis    & 330   & 187,790   & 288  & 6.84  & 2.08  & 5.73  & 1.83 & 38.9    & 289  & 6.54  & 2.09  & 4.90  & 1.99 & 47.6    \\
Montreal Notre Dame & 445   & 633,938   & 381  & 1.32  & 0.63  & 1.18  & 0.55 & 166.2   & 389  & 1.12  & 0.54  & 1.18  & 0.54 & 214.5   \\
Notre Dame           & 547   & 1,345,766 & 463  & 0.81  & 0.29  & 1.88  & 0.71 & 673.0   & 479  & 0.62  & 0.26  & 1.36  & 0.69 & 1,118.1 \\
NYC Library          & 313   & 259,302   & 267  & 6.14  & 1.89  & 2.42  & 1.21 & 39.5    & 263  & 6.67  & 2.25  & 2.52  & 1.28 & 42.1    \\
Piazza del Popolo   & 307   & 15,791    & 251  & 5.77  & 1.62  & 3.10  & 1.78 & 41.1    & 257  & 5.57  & 1.67  & 3.16  & 1.71 & 45.0    \\
Piccadilly            & 2,226 & 1,278,612 & 1980 & 3.46  & 1.69  & 3.17  & 1.80 & 1,287.4 & 1888 & 3.39  & 1.66  & 3.26  & 1.94 & 1,566.0 \\
Roman Forum          & 995   & 890,945   & 828  & 6.78  & 2.64  & 2.40  & 1.40 & 199.9   & 814  & 11.54 & 2.21  & 2.30  & 1.47 & 268.8   \\
Tower of London     & 440   & 474,171   & 364  & 21.15 & 6.06  & 3.23  & 2.36 & 43.8    & 359  & 27.39 & 10.53 & 3.98  & 3.57 & 48.8    \\
Union Square         & 733   & 323,933   & 612  & 12.18 & 7.60  & 6.90  & 3.20 & 39.1    & 610  & 13.90 & 8.60  & 11.64 & 4.20 & 53.0    \\
Vienna Cathedral     & 789   & 1,361,659 & 694  & 13.16 & 7.31  & 6.36  & 2.48 & 513.7   & 692  & 13.54 & 7.17  & 7.70  & 2.90 & 672.2   \\
Yorkminster           & 412   & 525,592   & 360  & 4.92  & 2.12  & 2.19  & 1.42 & 60.5    & 358  & 4.96  & 2.38  & 2.17  & 1.45 & 79.6    \\
Gendarmenmarkt        & 671   & 338,800   & 602  & 35.45 & 14.82 & 28.31 & 4.17 & 93.4    & 600  & 32.87 & 14.34 & 28.59 & 4.09 & 110.6  
\\
\bottomrule
\end{tabular}%
}
\caption{Performance of the LUD+TME and LUD+FMS pipelines \textbf{without matrix completion} on the Photo Tourism datasets: $n$ and $N$ are the number of cameras and key points, respectively (when applying screening, $n$ also denotes the number of the remaining cameras); $\hat{e}_R,\tilde{e}_R$ indicate mean and median errors of absolute camera rotations in degrees, respectively; $\hat{e}_T,\tilde{e}_T$ indicate mean and median errors of absolute camera translations in meters, respectively; $T_{\text{total}}$ is the runtime of the pipeline (in seconds). }
\label{tab:remove_camera_2_w/t_MC}
\end{table*}

\end{document}